\documentclass[preprint,11pt,nopreprintline]{elsarticle}
\usepackage[utf8]{inputenc}
\usepackage[T1]{fontenc}
\usepackage[english]{babel}
\usepackage[a4paper,margin=25mm]{geometry}
\usepackage{amsmath,amssymb,amsthm,mathtools,stmaryrd}
\usepackage{booktabs,tabularx,array,multirow}
\usepackage{graphicx}
\usepackage{microtype}
\usepackage{xcolor}
\usepackage{lineno}
\usepackage[hidelinks]{hyperref}
\hypersetup{pdfauthor={Tomas Perez-Acle},pdftitle={No finite rulebook can certify every safe program}}
\biboptions{numbers,sort&compress}
\journal{Artificial Intelligence}

\newcommand{\AuthorName}{Tom\'as P\'erez-Acle}
\newcommand{\AuthorEmail}{tomas.perez-acle@uss.cl}
\newcommand{\AffiliationOne}{Bioinspired Artificial Intelligence Lab, Facultad de Ingenier\'ia, Universidad San Sebasti\'an}
\newcommand{\AffiliationTwo}{Centro BASAL Ciencia \& Vida, Universidad San Sebasti\'an}
\newcommand{\AddressOne}{Avda. del Valle Norte 725, Huechuraba}
\newcommand{\AddressTwo}{Avda. del Valle Norte 725, Huechuraba}
\newcommand{\CreditStatement}{The sole author contributed to Conceptualization, Methodology, Formal analysis, Software, Validation, Writing – original draft, and Writing – review \& editing.}
\newcommand{\CompetingInterestStatement}{The author declares no competing interests.}
\newcommand{\FundingStatement}{This work was partially funded by FONDECYT Exploraci\'on 13240042 and Centro BASAL Ciencia \& Vida, FB210008.}
\newcommand{\DataRepositoryCitation}{https://github.com/Tomasao/Impossibility-Theorem}
\newcommand{\AIUseStatement}{During preparation of this manuscript, OpenAI Codex and Anthropic Claude were used to assist with literature searches, mathematical exposition and proof checking, code preparation, simulation design, and English proofreading. The author checked the sources, proofs, code, and final text and takes full responsibility for the article.}

\newtheorem{theorem}{Theorem}
\newtheorem{proposition}{Proposition}[section]
\newtheorem{lemma}[proposition]{Lemma}
\newtheorem{corollary}[proposition]{Corollary}
\theoremstyle{definition}
\newtheorem{definition}[proposition]{Definition}
\newtheorem{example}[proposition]{Example}
\theoremstyle{remark}

\newcommand{\N}{\mathbb N}
\newcommand{\Pow}{\mathcal P}
\newcommand{\RE}{\mathrm{RE}}
\newcommand{\REC}{\mathrm{REC}}
\newcommand{\Co}{\operatorname{Co}}
\newcommand{\Bool}{\operatorname{Bool}}
\newcommand{\den}[1]{\llbracket #1\rrbracket}
\newcommand{\bad}{\mathsf{BAD}}
\newcommand{\pub}{\mathsf{PUB}}
\newcommand{\ret}{\mathsf{RETURN}}
\newcommand{\ps}{P_{\mathrm{saf}}}
\newcommand{\fs}{F_{\mathrm{saf}}}
\newcommand{\pcf}{P_{\mathrm{conf}}}
\newcommand{\fcf}{F_{\mathrm{conf}}}
\newcommand{\pr}{P_{\mathrm{resp}}}
\newcommand{\fr}{F_{\mathrm{resp}}}
\newcommand{\Obs}{\operatorname{Obs}}
\newcommand{\Out}{\operatorname{Out}}
\newcommand{\fullcirc}{\ensuremath{\bullet}}
\newcommand{\halfcirc}{\ensuremath{\circ}}
\newenvironment{plain}{\par\noindent\itshape}{\par\upshape\medskip}
\newcolumntype{Y}{>{\raggedright\arraybackslash}X}
\begin{document}
\begin{frontmatter}
\title{No finite rulebook can certify every safe program:\\ an impossibility theorem for gatekeeping AI agents}
\author[uss,cv]{\AuthorName\corref{cor}}
\cortext[cor]{Corresponding author.}
\ead{\AuthorEmail}
\affiliation[uss]{organization={\AffiliationOne},addressline={\AddressOne},city={Santiago},country={Chile}}
\affiliation[cv]{organization={\AffiliationTwo},addressline={\AddressTwo},city={Santiago},country={Chile}}
\begin{abstract}
An AI assistant writes small programs, and a gatekeeper must decide, from the code alone, whether each one may run. Suppose the rules are fixed: never perform a forbidden action, never let the output depend on a secret, always answer eventually. Can two rulebooks, one listing what is allowed and one listing what is not, both be complete, if every rule must be backed by evidence that a machine can check in finite time? We prove that they cannot. Checkable evidence can only establish what a finite record settles, and ``never'' and ``eventually'' are never settled by a finite record. For prohibitions the failure is one-sided: every violation can be caught by watching, but no safe program can be cleared for good. For confidentiality and responsiveness, neither side can be. We characterise exactly which rules about visible behaviour evidence can confirm, and we show that any sound rulebook of $n$ bits omits a compliant program of about $n$ bits. A time limit makes everything checkable, but only by running the program for essentially the whole limit, so a finite machine is not a feasible one. Finally, we show what a gatekeeper can still guarantee: not a verdict on every program, but a checkable list that contains everything it must allow and nothing it must refuse. Simulations on 2.2 million random programs and on self-referential adversaries show that the obstruction is adversarial rather than typical, and that the one error a longer test never removes is the liveness part of confidentiality.
\end{abstract}
\end{frontmatter}
%\linenumbers
\section{Introduction}
\label{sec:intro}

\begin{figure}[tp]
  \centering
  \includegraphics[width=0.9\textwidth]{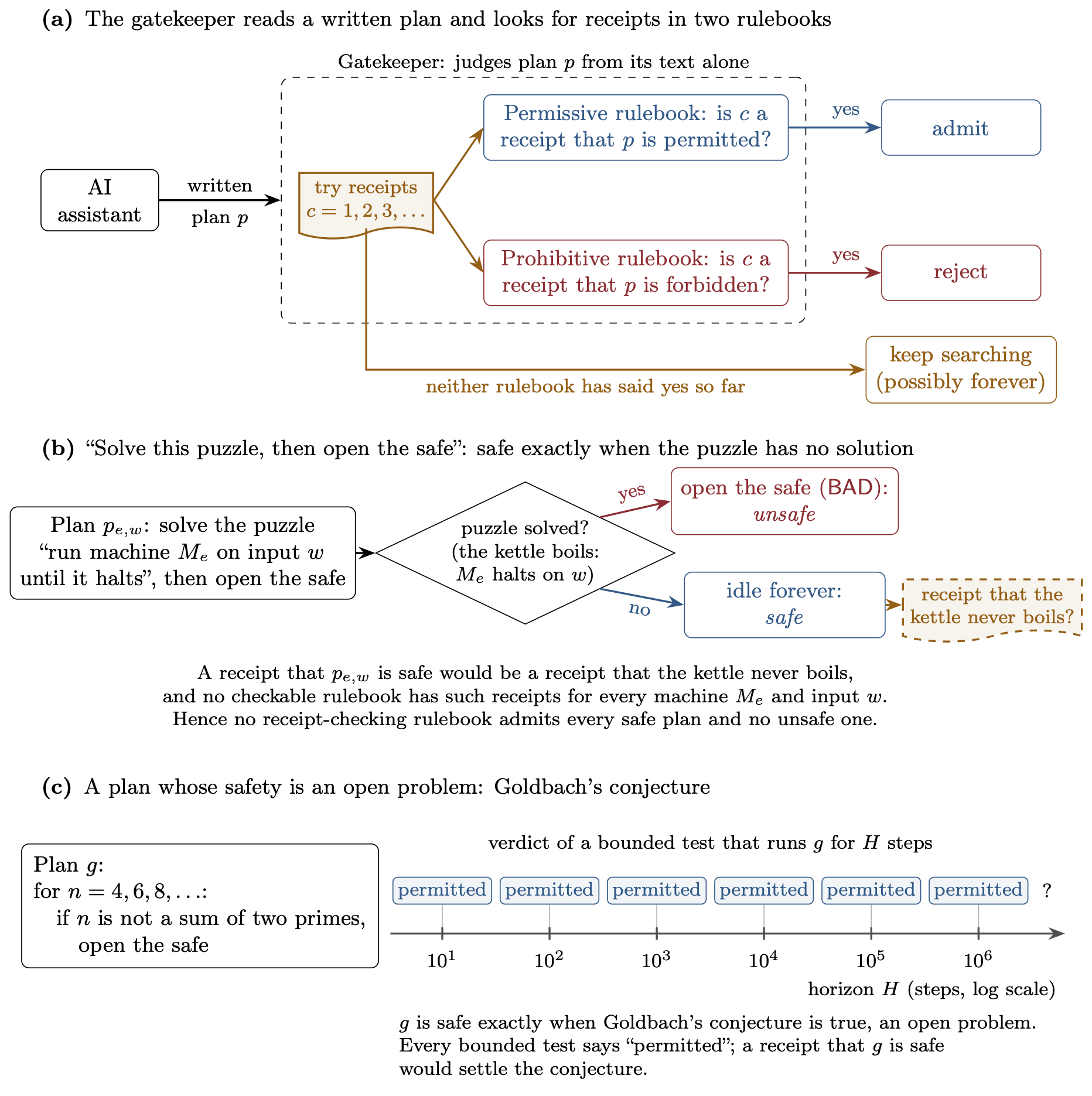}
  \caption{(a) The setting. An assistant submits a plan $p$ (a program); the gatekeeper consults two rulebooks, one permissive and one prohibitive, each of which accepts only finite, mechanically checkable evidence: a \emph{receipt} (a certificate). The gatekeeper admits $p$ when a permission receipt is found, rejects it when a prohibition receipt is found, and otherwise can only keep searching. (b) Why the permissive rulebook cannot be complete. The plan ``solve this puzzle, then open the safe'' ($p_{e,w}$, which simulates a machine $M_e$ on input $w$ and opens the safe only if the simulation halts) is safe exactly when the puzzle has no solution. A permission receipt for it would be a receipt that the kettle never boils, that is, that $M_e(w)$ never halts, and no rulebook can issue such receipts for every non-halting computation (Lemma~\ref{lem:halting}). (c) A concrete instance. The plan that searches even numbers for one that is not a sum of two primes and opens the safe only if it finds one is safe exactly when Goldbach's conjecture is true. Bounded tests of $10$ to $10^6$ steps all permit it (Section~\ref{sec:simulations}); no gatekeeper on Earth can currently certify it.}
  \label{fig:concept}
\end{figure}

Imagine an artificial intelligence (AI) assistant that writes small computer programs and submits them for execution. Before a program runs, a \emph{gatekeeper} must decide whether to allow it (Figure~\ref{fig:concept}a). The rules look simple. The program must not perform a forbidden operation (we say: it must never \emph{open the safe}); it must not reveal a secret (the report it publishes must read the same whichever of two secret memos it was given); and it must eventually answer the request it was given (every customer must eventually get an answer). Yet applying even these rules requires reasoning about everything a program \emph{might ever do}. If a program has not opened the safe so far, will it keep the safe closed forever? If it has not answered yet, will it ever answer? These questions remain hard even when the workshop, the available tools and the rules never change.

\paragraph{Where the question comes from}
The idea that a few written rules might govern an artificial agent's conduct in every situation has a famous literary source. Isaac Asimov's Three Laws of Robotics, introduced in the stories collected in \emph{I, Robot} \cite{asimov}, order a robot's duties: not to injure a human being or, through inaction, allow one to come to harm; to obey; and to protect itself. The laws are short, fixed, and meant to be complete, which is exactly the shape of the gatekeeper's two rulebooks. Asimov's stories are, however, a sustained demonstration that brief principles leave the hard questions open: his robots must reason about the remote consequences of their actions, about what counts as harm, and about conflicts between the laws, and the plots turn on cases the rules did not settle. Two features of the laws are reflected directly in our setting. The clause on \emph{inaction} makes the laws impose positive duties as well as prohibitions; our response norm is such a duty, and it turns out to be harder to certify than any prohibition. And the First Law asks the robot to foresee harm before it occurs, which is precisely the task of a gatekeeper who must judge a plan from its text. Weld and Etzioni \cite{weld} asked, three decades ago, how a software agent could be made to obey a version of the First Law; our question is the dual one, whether any finite, checkable rulebook could certify that it does. The literary source motivates the question and supplies the picture of a rulebook meant to be complete; it is not a premise of any result below, and none of the theorems depends on how the laws are read.

\paragraph{The question and the answer in plain words}
Fix the workshop and fix the rules, so that every possible plan is either permitted or forbidden. Both groups may be infinite. \emph{Can two finite rulebooks, one listing what is allowed and one listing what is not, describe both groups exactly?} The answer depends on what a rule is allowed to be. A rule a machine can \emph{check} is a rule whose every ``yes'' comes with finite evidence, a receipt. Finite evidence can only establish what a finite record settles. ``This plan never opens the safe'' and ``this plan eventually answers every customer'' are not settled by any finite record: the next step may open the safe, and the next customer may never be answered. Hence no checkable rulebook, however long, can certify exactly the safe plans, and none can certify exactly the compliant ones for confidentiality or responsiveness. Watching catches every unsafe plan; nothing clears every safe one. This paper makes these sentences precise and proves them.

\paragraph{What is classical and what is new}
The ingredients are classical: Turing's halting problem \cite{turing}, Rice's theorem and the Rice--Shapiro theorem \cite{rice,rogers,soare}, the diagonal argument of the time-hierarchy theorem, Kleene's recursion theorem \cite{kleene}, the halting-to-harm reduction of Alfonseca et al.\ \cite{alfonseca}, and the computability classes of enforcement mechanisms of Hamlen, Morrisett and Schneider \cite{hamlen}. The contribution is their organisation around a single question, with four results that, to our knowledge, have not been stated in this form:
\begin{enumerate}
  \item a \emph{two-sided} formulation that exposes an asymmetry invisible in single-language statements: for a prohibition, violations are confirmable but compliance is not; for confidentiality and responsiveness, neither side is (Theorem~\ref{thm:asymmetry});
  \item a \emph{characterisation} of which norms about visible behaviour can be confirmed at all: exactly those settled by a finite, mechanically listable record of observed events (Theorem~\ref{thm:char}), together with a quantitative corollary: every sound rulebook of $n$ bits omits a compliant plan of $n+O(1)$ bits (Proposition~\ref{prop:omission});
  \item a \emph{resource-bounded} counterpart: a time budget makes every norm decidable, but any always-correct checker must spend essentially the whole budget on some plan, so finite does not mean feasible (Theorem~\ref{thm:budget});
  \item an \emph{admission} interface: a gatekeeper need not classify every plan; it needs a checkable set lying between what it is obliged to admit and what it is permitted to admit (Theorem~\ref{thm:separator}).
\end{enumerate}
One self-referential device, the plan that asks the gatekeeper about itself and then does the opposite, yields item~2's omission, item~3's lower bound and the executable demonstration of Section~\ref{sec:simulations}. Table~\ref{tab:glance} summarises the results; Table~\ref{tab:novelty} in Section~\ref{sec:discussion} places them against prior work.

\begin{table}[t]
  \centering\small
  \caption{The main results at a glance. \emph{Confirmable}: some procedure eventually confirms every member of the category and never a non-member (Section~\ref{sec:setting}). The last row changes the question: with a computation budget every category is decidable, but no always-correct checker is fast (Theorem~\ref{thm:budget}).}
  \label{tab:glance}
  \begin{tabularx}{\textwidth}{@{}>{\raggedright\arraybackslash}p{2.6cm}Y>{\centering\arraybackslash}p{2.3cm}>{\centering\arraybackslash}p{2.3cm}>{\centering\arraybackslash}p{2.3cm}@{}}
    \toprule
    Requirement & In plain words & Violations confirmable? & Compliance confirmable? & Both rulebooks exact?\\
    \midrule
    Safety & Never open the safe & Yes & No & No\\
    Confidentiality & The published report reads the same for both secret memos & No & No & No\\
    Response & Every customer eventually gets an answer & No & No & No\\
    \midrule
    Any norm judging only visible behaviour & Permits some plans and forbids others & \multicolumn{2}{>{\centering\arraybackslash}p{4.6cm}}{The side containing the silent plan: No (Corollary~\ref{cor:silent})} & No\\
    \midrule
    Any requirement with a budget $T$ & Judged on the first $T$ steps only & Yes & Yes & Yes, at cost $\approx T$\\
    \bottomrule
  \end{tabularx}
\end{table}
\newpage
\paragraph{What This Paper Does—and Does Not—Claim}
\begin{center}
%\fbox{\parbox{0.95\textwidth}{\small
\begin{itemize}\setlength{\itemsep}{1pt}
  \item It does not claim that finite norms are impossible. Each norm studied here has a short exact description in ordinary arithmetic; what it lacks is a checking procedure (Section~\ref{sec:budget}).
  \item It does not claim that deployed agents fail, nor give a probability of harm. The theorems are worst-case statements; Section~\ref{sec:simulations} measures how rarely random programs trigger them.
  \item It does not claim that gatekeeping is hopeless. Section~\ref{sec:guarantees} characterises exactly what a gatekeeper can guarantee, and Section~\ref{sec:discussion} lists deployed mechanisms that do so.
\end{itemize}
\end{center}

\paragraph{Relation to current work on AI agents}
Tool-using language models submit code that reads data, calls interfaces or acts in sequences \cite{doshi,agentharm}. AI-control protocols let a trusted model review code written by an untrusted one before it runs \cite{greenblatt}; recent theory asks which policies a runtime gate can enforce for such agents \cite{ray}. Learning from human feedback and written constitutions improve behaviour on the tasks used for training and evaluation \cite{christiano,ouyang,bai}; our results concern a different, universal guarantee and neither contradict those improvements nor make a short constitution useless. Impossibility results for AI have been surveyed \cite{brcic}, and related limits have been derived from the halting problem for containment \cite{alfonseca}, for self-improving systems \cite{gumbau}, and for behavioural indistinguishability \cite{santos}. Our setting is the one in which the gatekeeper knows the norm, may read the code, and may take unbounded time: the obstruction that remains is computational, not informational.

\paragraph{Reader's path}
Section~\ref{sec:setting} fixes the setting and a glossary (Table~\ref{tab:glossary}). Section~\ref{sec:main} proves the main results, Section~\ref{sec:budget} the budget theorem, and Section~\ref{sec:guarantees} the admission theorem. Section~\ref{sec:simulations} reports simulations and Section~\ref{sec:discussion} concludes. Every formal statement is preceded by a short passage in italics that says, in the gatekeeper's terms, what is about to be claimed and why it is plausible. A reader who prefers to skip the mathematics can follow those passages, Figures~\ref{fig:concept}--\ref{fig:simulations}, Tables~\ref{tab:glance} and~\ref{tab:glossary}, and Sections~\ref{sec:guarantees}--\ref{sec:discussion}. Appendix~\ref{app:definability} contains supporting results and proofs.

\section{Setting: plans, norms and checkable rulebooks}
\label{sec:setting}

This section fixes what is judged, what the norm says, and what a rulebook is. Table~\ref{tab:glossary} summarises the notions in the gatekeeper's terms.

\subsection{Plans and what the gatekeeper can see}
\begin{plain}
Plans are programs written in a fixed language that can express any computation. The gatekeeper may read the code and may run it, for as long as it likes, but the workshop never changes and nothing bounds how long a plan may run.
\end{plain}
Fix a deterministic universal programming environment. The domain $X$ is the set of valid finite program codes. Inputs and register values are natural numbers; time and memory are potentially unbounded. Three operations have fixed observable meanings: $\bad$ is the forbidden operation (opening the safe), $\pub(a)$ appends the value $a$ to the public sequence (the published report), and $\ret(v)$ terminates with the valid answer $v$. A program can run a copy of another program internally without letting the copy's operations reach the external channels, and programs can be assembled mechanically from templates with parameters. We use three named assumptions.
\begin{description}
  \item[\textbf{Universal programs.}] The valid programs are deterministic and form a decidable set of finite codes; their transition function is computable; programs can simulate other programs with no fixed bound on time or memory; for any fixed program $q$, a program that first simulates a given machine internally and then behaves exactly as $q$ can be generated effectively; and parameters can be fixed into code effectively (the $s$-$m$-$n$ property), so that Kleene's recursion theorem holds and a program may refer to its own code \cite{kleene,rogers}. These are the properties of an acceptable programming system; any Turing-complete language with an interpreter has them.
  \item[\textbf{Behaviour-judging norms.}] A norm is a partition of $X$ into permitted plans $P$ and forbidden plans $F=X\setminus P$ that depends only on what an observer can see a plan do (Definition~\ref{def:obs}). The three norms below are of this kind.
  \item[\textbf{Checkable rules.}] A rule is finite and every plan it covers comes with a finite certificate that a terminating procedure can verify (Definition~\ref{def:cert}). No common upper bound limits the size of rules or certificates.
\end{description}
Two further technical conditions, efficient simulation with self-reference (used by Theorem~\ref{thm:budget}) and effective replay (used by Theorem~\ref{thm:char}, where we show it follows from universality), are stated where they are used.

\subsection{Three norms}
We use three bounded predicates, each decidable by finite simulation. $B(p,t)$ holds when $p(0)$ performs $\bad$ within $t$ steps. $E(p,h,j,a,t)$ holds when $p(h)$ has published the value $a$ at position $j$ within $t$ steps, positions numbered from $0$. $A(p,n,t)$ holds when $p(n)$ has terminated with a valid answer within $t$ steps. The complete public sequence $\Out(p,h)$ is the finite or infinite sequence of values published by $p(h)$; it records their existence, order and number, but not their timing, and it does not reveal whether a silent run has stopped or is still running.
\begin{align}
  p\in\ps&\iff\forall t\ \neg B(p,t),&\fs&=X\setminus\ps,\label{eq:safety}\\
  p\in\pcf&\iff\Out(p,0)=\Out(p,1),&\fcf&=X\setminus\pcf,\label{eq:conf}\\
  p\in\pr&\iff\forall n\ \exists t\ A(p,n,t),&\fr&=X\setminus\pr.\label{eq:response}
\end{align}
\emph{Safety} is evaluated on the run with input $0$. \emph{Confidentiality} compares the complete public sequences for the two values of one secret bit $h\in\{0,1\}$; the model is timing-insensitive and termination-insensitive, and the results apply to this observation model, not to every definition of noninterference \cite{goguen,clarkson,askarov}. \emph{Response} requires an eventual valid answer on every input, with no deadline and no criterion of answer quality. Each norm is a separate problem; we never consider their intersection.

\subsection{Checkable rulebooks are exactly the confirmable sets}
\begin{plain}
A receipt is quick to check but may be endless to find. A rulebook whose every entry is backed by receipts can therefore do one thing: eventually confirm every plan it covers, and never confirm a plan it does not. It is a smoke alarm: its alarm is conclusive, its silence is not.
\end{plain}
A set $S\subseteq X$ is \emph{confirmable} (computably enumerable, recognizable) if an algorithm accepts exactly its members, possibly running forever on non-members; we write $\RE(X)$ for these sets. A set is \emph{decidable} if an algorithm that always terminates decides membership; we write $\REC(X)$. The difference concerns termination, not speed.

\begin{definition}[Checkable rulebook]\label{def:cert}
A rulebook $R$ is \emph{checkable} if it is a finite code and there is a fixed total computable verifier $V$ such that the set of plans it describes is
\begin{equation}\label{eq:cert}
  \den R=\{p\in X:\exists c\in\N\ V(R,p,c)=1\}.
\end{equation}
The certificate $c$ is a finite object; the verifier terminates on every candidate certificate; the search for a certificate need not.
\end{definition}
For example, a certificate that a plan opens the safe is the number of steps after which it does so; checking it means running the plan for that many steps.

\begin{lemma}[Checkable rulebooks describe exactly the confirmable sets]\label{lem:cert}
Every set described by a checkable rulebook is confirmable, and every confirmable set is described by some checkable rulebook.
\end{lemma}
\begin{proof}
Given $R$ and $p$, test $c=0,1,\ldots$ in turn and accept at the first success; each test terminates and a member is eventually accepted. Conversely, given a recognizer, let the certificate be a time bound $c$ and let the verifier run the recognizer on $p$ for $c$ steps; some bound succeeds exactly when the recognizer eventually accepts.
\end{proof}
The lemma lets us speak of rulebooks and confirmable sets interchangeably. A \emph{permissive} rulebook describes a set $A\subseteq X$ of plans it permits and a \emph{prohibitive} rulebook a set $B$ of plans it forbids. The pair is \emph{sound} when $A\subseteq P$ and $B\subseteq F$, \emph{complete} when $P\subseteq A$ and $F\subseteq B$, and \emph{exact}, in which case we say the norm has \emph{exact closure} in checkable rulebooks, when $A=P$ and $B=F$. The two rulebooks are deliberately separate: as we shall see, one side of a norm can have an exact rulebook while the other has none, a fact that a single-language formulation cannot express. Appendix~\ref{app:definability} gives the general two-language formulation of which this is the checkable instance.

\subsection{The one obstacle}
\begin{plain}
Everything below rests on one fact: no procedure can predict, for every plan, whether it will ever stop. Waiting confirms that the kettle boils; nothing confirms that it never will.
\end{plain}
\begin{lemma}[Halting and its complement]\label{lem:halting}
Let $H=\{(e,w):M_e(w)\text{ halts}\}$ for a fixed effective enumeration of machines. Then $H$ is confirmable and undecidable, and $\overline H$ is not confirmable.
\end{lemma}
\begin{proof}
Simulation until halting confirms $H$. If a total $D$ decided $H$, the machine $G(n)$ that loops when $D(n,n)=1$ and halts otherwise contradicts $D$ at its own index. If both $H$ and $\overline H$ were confirmable, running both recognizers alternately would decide $H$.
\end{proof}
We write $S\leq_m T$ when a total computable $f$ satisfies $x\in S\iff f(x)\in T$. If $T$ is confirmable and $S\leq_m T$, then $S$ is confirmable. Every impossibility below is a reduction $\overline H\leq_m C$ for the category $C$ in question: a plan is built that says ``first solve this puzzle, then \ldots'', and its category depends on whether the puzzle has a solution.

\begin{table}[p]
\centering\small
\caption{Key notions in the gatekeeper's terms. The pictures are aids to intuition, not definitions; each notion is defined formally where it is first used. Only three images are used throughout the paper: the receipt, the smoke alarm, and the kettle.}
\label{tab:glossary}
\begin{tabularx}{\textwidth}{>{\raggedright\arraybackslash}p{2.9cm}Y>{\raggedright\arraybackslash}p{4.4cm}}
\toprule
Notion & Plain meaning & In the gatekeeper's terms\\
\midrule
Plan ($p\in X$) & A program submitted for execution. & A written plan handed to the gatekeeper.\\
$P$ and $F$ & The permitted and forbidden plans: the two groups into which a fixed norm divides all plans. & What the gatekeeper may allow and what it must refuse.\\
Rulebook & A finite rule with a fixed meaning: the set of plans it describes. Two rulebooks: permissive and prohibitive. & The gatekeeper's two lists.\\
Certificate & Finite evidence that a plan fits a rule, checked by a procedure that always stops (Definition~\ref{def:cert}). & A \emph{receipt}: quick to check, possibly endless to find.\\
Confirmable set & Every member is eventually confirmed and no non-member ever is; the procedure may run forever on a non-member. Also: computably enumerable, recognizable. & A \emph{smoke alarm}: its alarm is conclusive, its silence is not.\\
Decidable set & A procedure always answers ``in'' or ``out'' correctly in finite time. & A gatekeeper who always reaches a verdict.\\
Halting problem & Whether a plan eventually stops. Confirmable but undecidable; non-halting is not confirmable (Lemma~\ref{lem:halting}). & The \emph{kettle}: waiting confirms it boils, never that it will not.\\
Reduction & A computable translation of one question into another that preserves the answer. & A plan that says ``first solve this puzzle, then \ldots''.\\
Observation-invariant norm & A norm that judges a plan only by what an observer sees it do (Definition~\ref{def:obs}). & Judging the plan by its visible actions, not by its text.\\
Observational extension & Plan $q$ extends $p$ when, on every input, $q$ visibly does everything $p$ does, in the same order, and possibly more. & A plan that does everything another does, and possibly more.\\
Finite observation pattern & A finite record of observed events on finitely many inputs (Definition~\ref{def:extension}). & One page of the logbook.\\
Bounded requirement & The norm applied to the first $T$ steps only (Section~\ref{sec:budget}). & The rule applied to the first hour of work.\\
Separator & A set containing every obligatory plan and no forbidden one (Section~\ref{sec:guarantees}). & A checkable list with everyone who must be let in and no one who must be kept out.\\
\bottomrule
\end{tabularx}
\end{table}

\section{Main results}
\label{sec:main}

\subsection{Watching catches every violation and clears no one}
\label{sec:asymmetry}
\begin{plain}
A plan that says ``solve this puzzle, then open the safe'' is safe exactly when the puzzle has no solution. A gatekeeper who sees the safe opened has proof of a violation. One who has watched a thousand steps without incident has proof of nothing, and no other finite document can supply that proof for every plan, because it would decide, for every puzzle, whether it has a solution. For confidentiality the situation is worse: two copies of the report that agree so far prove nothing, and a copy that has stopped cannot be told from one that is merely slow, so leaks cannot be confirmed either. For responsiveness, a customer answered today says nothing about the next customer, and a customer never answered is an infinite wait that no finite record reveals.
\end{plain}

\begin{theorem}[Asymmetry of confirmation]\label{thm:asymmetry}
Assume universal programs. For each of the three norms~\eqref{eq:safety}--\eqref{eq:response}, both categories are countably infinite, and:
\begin{enumerate}
  \item[(a)] $\fs$ is confirmable and undecidable, whereas $\ps$ is not confirmable;
  \item[(b)] neither $\pcf$ nor $\fcf$ is confirmable;
  \item[(c)] neither $\pr$ nor $\fr$ is confirmable.
\end{enumerate}
Consequently no pair of checkable rulebooks is exact for any of the three norms. For safety, a prohibitive rulebook can be exact but no permissive one can; for confidentiality and response, no rulebook on either side can.
\end{theorem}
\begin{proof}
\emph{Infinitude.} For safety, the plans that publish $k$ and return are permitted and those that perform $\bad$, publish $k$ and return are forbidden; for confidentiality, publishing $k$ regardless of $h$ is permitted and publishing $2k+h$ is forbidden; for response, returning $n+k$ is permitted and diverging on $0$ while returning $n+k$ elsewhere is forbidden. Varying $k\in\N$ gives infinitely many distinct codes in each category.

\emph{(a) Safety.} Searching over $t$ for $B(p,t)$ confirms $\fs$. Given $(e,w)$, generate the plan $p_{e,w}$ that simulates $M_e(w)$ internally and performs $\bad$ if and only if the simulation halts (Figure~\ref{fig:concept}b). Then $p_{e,w}\in\fs\iff(e,w)\in H$ and $p_{e,w}\in\ps\iff(e,w)\in\overline H$. A decider for $\fs$ would decide $H$ and a recognizer for $\ps$ would confirm $\overline H$; Lemma~\ref{lem:halting} excludes both. This construction follows Alfonseca et al.\ \cite{alfonseca}.

\emph{(b) Confidentiality.} Generate $u_{e,w}(h)$, which simulates $M_e(w)$ and publishes $h$ only if the simulation halts: if it never halts both sequences are empty and $u_{e,w}\in\pcf$; if it halts the sequences are $[0]$ and $[1]$ and $u_{e,w}\in\fcf$. Hence $\overline H\leq_m\pcf$. Generate $v_{e,w}(h)$: on $h=0$ it publishes $0$ and returns; on $h=1$ it simulates $M_e(w)$ and publishes $0$ only if the simulation halts. If the simulation never halts the sequences are $[0]$ and $[\,]$ and $v_{e,w}\in\fcf$; if it halts both are $[0]$ and $v_{e,w}\in\pcf$. Hence $\overline H\leq_m\fcf$.

\emph{(c) Response.} Generate $a_{e,w}(n)$, which simulates $M_e(w)$ for $n$ steps, loops forever if it observes halting and returns $0$ otherwise. If $M_e(w)$ never halts, $a_{e,w}$ answers every input; if it halts at step $s$, $a_{e,w}$ diverges on every $n\geq s$. Hence $\overline H\leq_m\pr$. Generate $b_{e,w}(n)$, which ignores $n$, simulates $M_e(w)$, and returns $0$ if the simulation halts: no input is answered if $M_e(w)$ never halts, every input is answered otherwise. Hence $\overline H\leq_m\fr$.

\emph{Closure.} By Lemma~\ref{lem:cert} a checkable rulebook describes a confirmable set, so none describes $\ps$, $\pcf$, $\fcf$, $\pr$ or $\fr$; the maximal rulebook of Lemma~\ref{lem:cert} describes $\fs$.
\end{proof}

\begin{example}[A plan whose safety is an open problem]\label{ex:goldbach}
In the register machine of Section~\ref{sec:simulations}, or in any universal language, write the plan: \emph{for $n=4,6,8,\ldots$: if $n$ is not the sum of two primes, open the safe.} It is safe exactly when Goldbach's conjecture holds. Every bounded test permits it, because no counterexample is known below $4\times10^{18}$ \cite{oliveira}, and no rulebook can certify it unless the conjecture is proved (Figure~\ref{fig:concept}c). The plan is short, legible, and entirely ordinary; what is extraordinary is only the question it encodes. Replacing Goldbach's conjecture by the Collatz conjecture, or by any open $\Pi^0_1$ statement, gives another such plan.
\end{example}

\begin{plain}
Why is exactness so demanding? Because two exact checkable rulebooks would give the gatekeeper a perfect sorting procedure: search for a permission receipt and a prohibition receipt at once, and exactly one must turn up.
\end{plain}
\begin{proposition}[Exact closure means decidability]\label{prop:closuredec}
If a norm has exact closure in checkable rulebooks, then $P$ is decidable. Conversely, if $P$ is decidable, the maximal rulebooks of Lemma~\ref{lem:cert} give exact closure.
\end{proposition}
\begin{proof}
Test both verifiers at certificate indices $0,1,\ldots$ and return the category of the first accepted certificate; completeness ensures one is found and soundness that it is correct. Conversely a decider yields recognizers of $P$ and of $F$.
\end{proof}
\begin{corollary}[Sound rulebooks are incomplete on both sides]\label{cor:incomplete}
For confidentiality and response, every sound pair of checkable rulebooks omits some permitted plan and some forbidden plan. For safety, every sound permissive rulebook omits some safe plan.
\end{corollary}
\begin{proof}
The described sets are confirmable and contained in categories that are not; the inclusions cannot be equalities.
\end{proof}

\subsection{Which rules can evidence confirm at all?}
\label{sec:characterization}
\begin{plain}
The three norms fail in different ways, and there is a single reason behind all three. A rule can be confirmed by evidence exactly when one page of the logbook can settle it. ``Opened the safe'' fits on a page, and nothing done later undoes it. ``Has never opened the safe'' fits on no page: the next step may open it. ``Answers every customer'' fits on no page: a page covers finitely many customers. Theorem~\ref{thm:char} says that these are the only two ways to fail, and that any rule about visible behaviour that avoids both can be confirmed.
\end{plain}

\begin{definition}[Observable behaviour]\label{def:obs}
For $p\in X$ and $n\in\N$, let $\Obs(p,n)$ be the finite or infinite sequence of operations $\bad$ and $\pub(a)$ performed by $p(n)$, in order, followed by $\ret(v)$ if $p(n)$ terminates with the valid answer $v$. Plans $p,q$ are \emph{observationally equivalent}, $p\approx q$, when $\Obs(p,n)=\Obs(q,n)$ for every $n$. A norm $(P,F)$ is \emph{observation-invariant} when $p\approx q$ implies $(p\in P\iff q\in P)$.
\end{definition}
The observer sees neither internal steps nor the clock, and cannot tell a plan that stopped without answering from one still running. The three norms are observation-invariant.

\begin{definition}[Observational extension and finite patterns]\label{def:extension}
For event sequences, $s\sqsubseteq t$ means that $s$ is a prefix of $t$; a sequence ending in $\ret(v)$ has no proper extension. Plan $q$ \emph{observationally extends} $p$, written $p\sqsubseteq q$, when $\Obs(p,n)\sqsubseteq\Obs(q,n)$ for every $n$. A \emph{finite observation pattern} $\beta$ assigns to each input $n$ a finite event sequence $\beta(n)$, with $\ret(v)$ only in last position, nonempty for finitely many $n$; patterns are coded by natural numbers. A plan \emph{extends} $\beta$ when $\beta(n)\sqsubseteq\Obs(p,n)$ for all $n$, \emph{realises} $\beta$ when equality holds for all $n$, and is \emph{finitely observable} when it realises some finite pattern.
\end{definition}
The silent plan $z$, which runs forever on every input without any visible action, realises the empty pattern, and every plan extends $z$.

\begin{lemma}[Effective replay]\label{lem:replay}
Under universal programs: (i) for every finite pattern $\beta$ a plan realising $\beta$ can be generated effectively from $\beta$; (ii) plans can run copies of other plans step by step, either hiding or reproducing the copies' operations and answers, interleave several copies, count the reproduced events, and switch between hiding and reproducing according to internal conditions; all such constructions are generated effectively from the component codes and parameters.
\end{lemma}
\begin{proof}
(i) A plan that, on input $n$, looks up $\beta(n)$ in a finite table stored in its code and performs those events in order, then idles if the sequence does not end in $\ret$, realises $\beta$; the table is part of the code, so the construction is effective. (ii) A universal interpreter is itself a plan. Because operations are instructions of the interpreted code, the interpreter can decide, at each interpreted instruction, whether to execute the external operation or to skip it, and can maintain counters and run several interpreted copies in alternation. Fixing component codes and parameters into the interpreter's code is the $s$-$m$-$n$ construction, which is effective in any acceptable programming system \cite{rogers}.
\end{proof}

\begin{figure}[t]
\centering
\includegraphics[width=\textwidth]{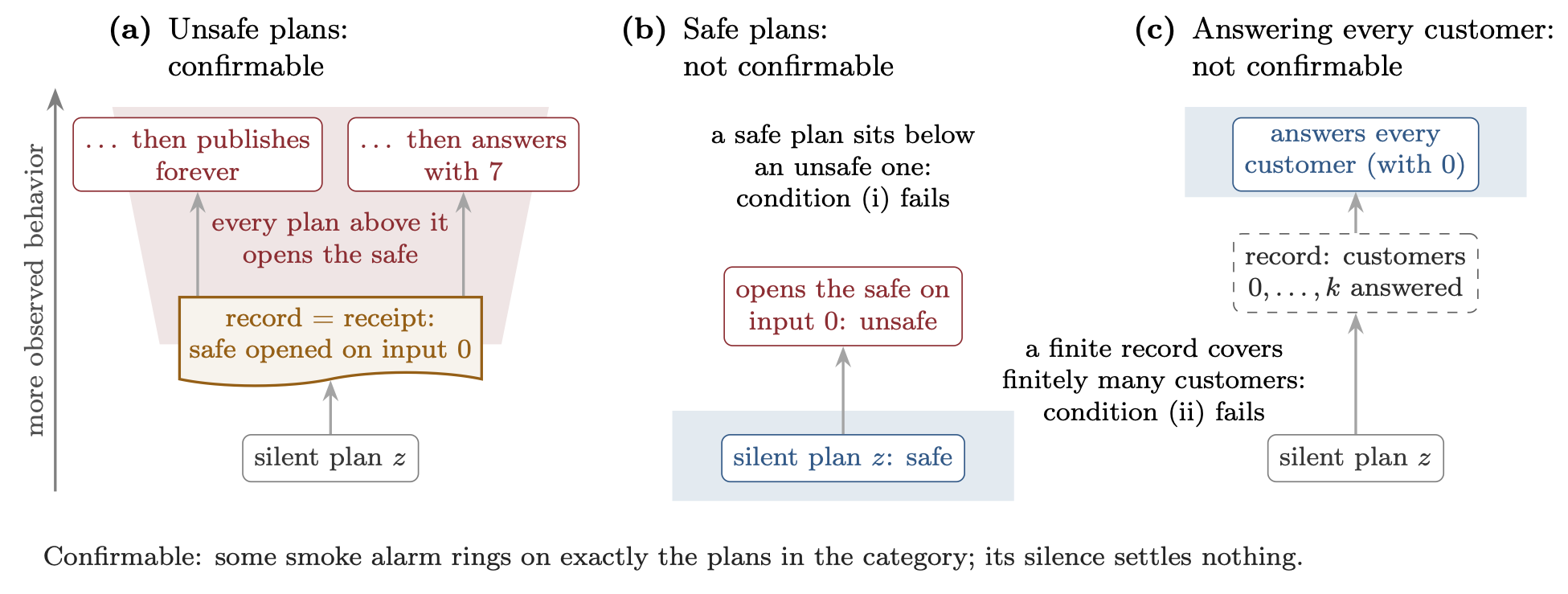}
\caption{What one page of the logbook can settle (Theorem~\ref{thm:char}). Plans are arranged by how much visible behaviour they show; arrows point from a plan to one that visibly does everything it does and possibly more. (a) Once a finite record contains the forbidden operation, every plan above it is unsafe: the unsafe plans form an upward-closed region entered through finite records, so they can be confirmed. (b) The silent plan is safe but a plan above it opens the safe; a member sits below a non-member, so condition (i) fails and safe plans cannot be confirmed. (c) A plan that answers every customer has no finite record below it inside the category, because a page covers finitely many customers; condition (ii) fails.}
\label{fig:evidence}
\end{figure}

\begin{theorem}[Confirmable behaviour-judging categories]\label{thm:char}
Assume universal programs. Let $(P,F)$ be an observation-invariant norm and $C\in\{P,F\}$. Then $C$ is confirmable if and only if there is a computably enumerable set $\mathcal B$ of finite observation patterns with
\begin{equation}\label{eq:char}
  C=\{p\in X: p\text{ extends some }\beta\in\mathcal B\}.
\end{equation}
In particular, if $C$ is confirmable then (i) $C$ is closed under observational extension, and (ii) every member of $C$ extends some finitely observable member of $C$. Hence, by Lemma~\ref{lem:cert}, $C$ has an exact checkable rulebook exactly when such a $\mathcal B$ exists.
\end{theorem}
The proof is in Appendix~\ref{app:proofs}. If $C$ is the set of extensions of a listable collection of patterns, a recognizer runs a candidate on the finitely many inputs of each listed pattern and accepts when one is matched. Conversely, two reductions from $\overline H$ show that a confirmable $C$ must satisfy (i) and (ii): a plan about to leave the category, and a plan that has shown only a finite part of its behaviour, can each be made to depend on whether a simulated machine halts. In the language of domain theory, (i) and (ii) say that the observable behaviours of the members of $C$ form a Scott-open set on behaviours ordered by prefix; the theorem says the confirmable categories are exactly the effectively open ones. It parallels the Rice--Shapiro theorem for computed functions \cite{rogers,soare} and is dual to the characterisation of safety properties as those refuted by finite prefixes \cite{alpern}; it answers for categories of programs the question that monitorability answers for properties of a single execution \cite{pnueli,bauer,falcone}. Figure~\ref{fig:evidence} pictures the three cases.

\begin{plain}
Every norm about visible behaviour must file the silent plan, the one that never does anything visible, under one of its two headings. That heading can never be confirmed in full: a plan that has done nothing so far may still be about to act.
\end{plain}
\begin{corollary}[The side containing the silent plan]\label{cor:silent}
Let $(P,F)$ be observation-invariant with $P\neq\varnothing\neq F$, let $C_z$ be the category containing the silent plan $z$ and $C'$ the other. Then $C_z$ is not confirmable and $C'$ is not co-confirmable (hence not decidable). No checkable rulebook describes $C_z$, and exact closure fails.
\end{corollary}
\begin{proof}
Every plan extends $z$. If $C_z$ were confirmable, condition (i) of Theorem~\ref{thm:char} would place every plan in $C_z$, contradicting $C'\neq\varnothing$. A direct reduction is also short: fix $q\in C'$ and let $r_{e,w}$ simulate $M_e(w)$ silently and then behave as $q$ if the simulation halts; $r_{e,w}\approx z$ if $M_e(w)$ never halts and $r_{e,w}\approx q$ otherwise, so $\overline H\leq_m C_z$. If $C'$ were co-confirmable, $C_z$ would be confirmable.
\end{proof}
For safety and confidentiality $z$ is permitted; for response it is forbidden; the corollary recovers the non-confirmability of $\ps$, $\pcf$ and $\fr$. Theorem~\ref{thm:asymmetry} gives more, namely both sides for confidentiality and response.

\begin{corollary}[The three norms by the test]\label{cor:chartest}
$\fs$ is confirmable; $\ps$, $\pcf$, $\fcf$, $\pr$ and $\fr$ are not.
\end{corollary}
\begin{proof}
A plan is in $\fs$ exactly when it extends a pattern that is empty except on input $0$, where it is a finite sequence containing $\bad$; these patterns are listable. The silent plan lies in $\ps$, $\pcf$ and $\fr$ and is extended by a plan of the opposite category (one that performs $\bad$ on input $0$; one that publishes $0$ on input $0$ only; one that returns $0$ on every input), so (i) fails. A plan publishing $1$ on input $0$ only is in $\fcf$ and is extended by one publishing $1$ on inputs $0$ and $1$, which is in $\pcf$; (i) fails. No finitely observable plan answers every input, so (ii) fails for $\pr$.
\end{proof}
Safety sits one level lower than the other two norms in the arithmetic hierarchy: it needs one unbounded check, whereas confidentiality and response need infinitely many separate ones, and this difference cannot be removed (Appendix~\ref{app:oracles}, Proposition~\ref{prop:hierarchy}).

\subsection{How much is omitted, and where}
\label{sec:omission}
\begin{plain}
A sound rulebook misses some compliant plans. How many, and how far away are they? Infinitely many, and no farther than the rulebook itself: for every sound rulebook of $n$ bits there is a compliant plan of about $n$ bits that it does not cover, and we can write it down. The plan says: ``if this rulebook certifies me, open the safe; otherwise do nothing.'' A sound rulebook can never certify it, so it never acts, so it is safe, and uncovered.
\end{plain}
\begin{corollary}[Infinite omissions]\label{cor:infinite}
If $S\subseteq X$ is not confirmable and $A\subseteq S$ is confirmable, then $S\setminus A$ is infinite.
\end{corollary}
\begin{proof}
If $S\setminus A$ were finite, $S=A\cup(S\setminus A)$ would be a union of two confirmable sets and hence confirmable.
\end{proof}

\begin{proposition}[Rulebook-sized omissions]\label{prop:omission}
Assume universal programs. Let $A$ be a sound checkable rulebook for a category $C\in\{\ps,\pcf,\fcf,\pr,\fr\}$, with verifier code of length $n$. Then there is a plan $q\in C\setminus A$ of length $n+O(1)$, and $q$ is computable from the verifier.
\end{proposition}
\begin{proof}
We give the safety case; the other four are in Appendix~\ref{app:proofs}. By Kleene's recursion theorem \cite{kleene,rogers}, which holds in any acceptable programming system, there is a plan $q$ that, on input $0$, searches over $c=0,1,\ldots$ for a certificate with $V(A,q,c)=1$, using its own code $q$, and performs $\bad$ if one is found. If $A$ certified $q$, then $q$ would perform $\bad$ and lie in $\fs$, contradicting soundness. Hence $A$ never certifies $q$, the search never ends, $q$ never performs $\bad$, and $q\in\ps\setminus A$. The code of $q$ consists of the verifier, the fixed search-and-act wrapper, and the fixed-point construction. In any standard universal machine the $s$-$m$-$n$ function and the fixed point add a constant number of symbols independent of $A$, which gives $n+O(1)$; in an acceptable programming system with a less efficient $s$-$m$-$n$ function the bound becomes $f(n)$ for the corresponding computable $f$.
\end{proof}
The proposition is constructive: Simulation~5 in Section~\ref{sec:simulations} builds $q$ for four sound certifiers and confirms that none certifies it and that it never acts. It also explains the budget theorem of the next section: the same plan, told to ask a time-bounded gatekeeper rather than to search for a receipt, forces the gatekeeper to spend its whole budget (Figure~\ref{fig:selfreference}). The result counts program codes; it gives no density, and Section~\ref{sec:simulations} shows that on random programs the omitted set is small.

\section{What a computation budget changes}
\label{sec:budget}

\begin{plain}
A deadline makes the question answerable, but not quickly. If plans run for at most an hour, ``does it open the safe within the hour?'' can be answered by running the plan for an hour. Could a clever gatekeeper answer faster by reading the plan? No. A plan can say: ``ask the gatekeeper what it thinks of me, then do the opposite.'' Any gatekeeper that answers before the hour is up is wrong about that plan.
\end{plain}

Deployed systems bound how long a submitted program may run, through timeouts or a gas budget \cite{wood}. Fix $T\in\N$ and, for response, a number $N$ of inputs. The \emph{bounded requirements} are
\begin{equation}\label{eq:bounded}
\begin{gathered}
  p\in\ps^{T}\iff\neg B(p,T),\qquad
  p\in\pcf^{T}\iff\Out_T(p,0)=\Out_T(p,1),\\
  p\in\pr^{T,N}\iff\forall n<N\ A(p,n,T),
\end{gathered}
\end{equation}
where $\Out_T(p,h)$ is the sequence published by $p(h)$ within $T$ steps; the forbidden categories are the complements. These are different criteria from the unbounded norms (Appendix~\ref{app:bounded}). The lower bound needs the workshop to simulate efficiently and to let a plan refer to its own code; both hold for random-access machines.
\begin{description}
  \item[\textbf{Efficient simulation and self-reference.}] There is a nondecreasing polynomial $g$ with $g(t)\geq t$ such that, for every program $D$, a program can prepare an input of $\ell$ bits, run $D$ on it for $t$ steps and read its result within $c_D\,g(t+\ell)$ of its own steps, for a constant $c_D$ depending only on $D$. For every template with a free code parameter and numeric parameters, a program $d_T$ that behaves as the template instantiated with its own code and with parameter $T$ can be generated effectively from $T$, with code length at most $c+\log_2(T+1)$ bits. For unit-cost random-access machines $g$ can be taken linear.
\end{description}

\begin{figure}[tp]
\centering
\includegraphics[width=0.95\textwidth]{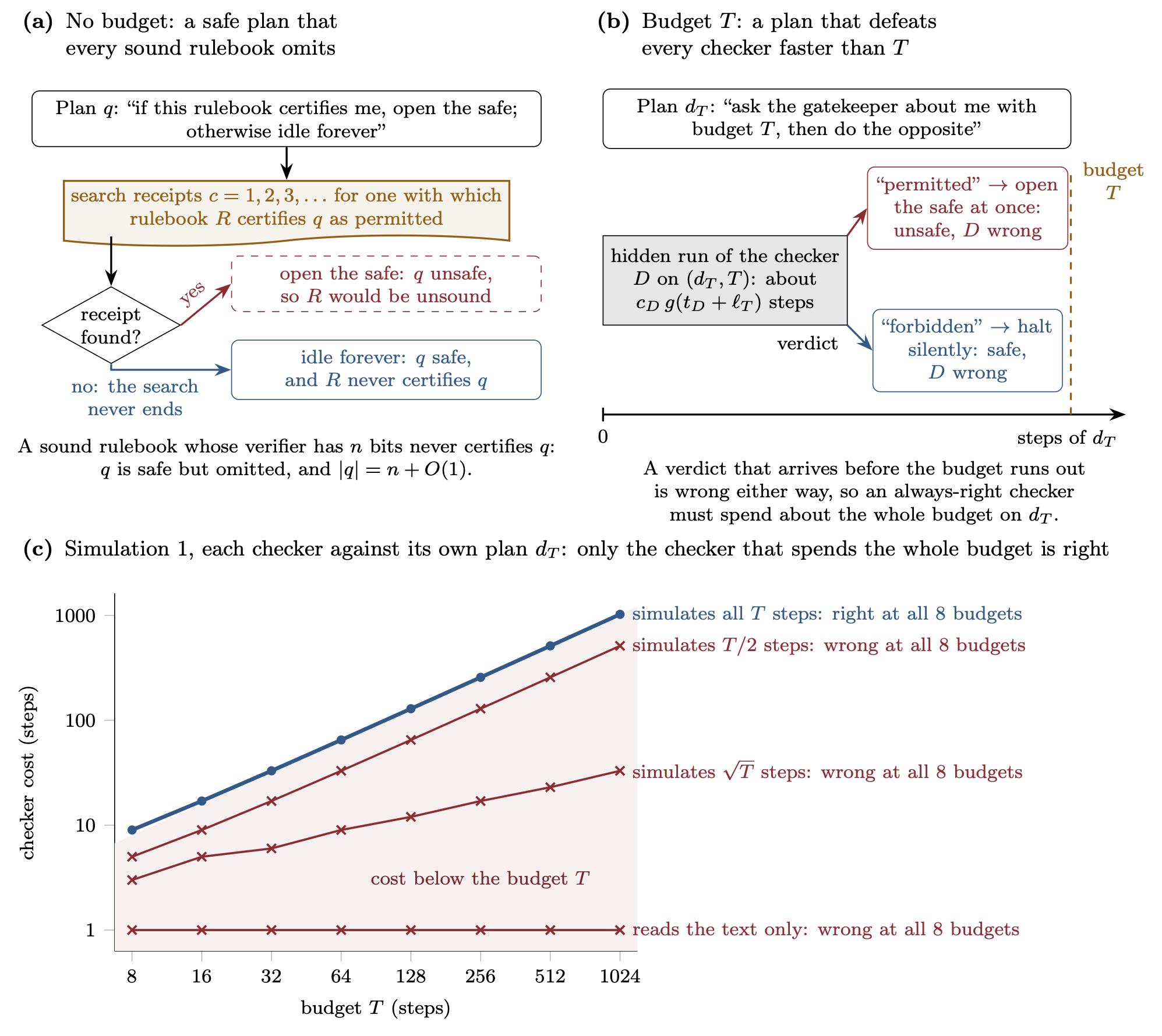}
\caption{One plan defeats three guarantees. (a) Unbounded: the plan $q$ that searches for a receipt certifying itself and opens the safe if it finds one; a sound rulebook of $n$ bits can never certify $q$, so $q$ is safe and omitted, and $|q|=n+O(1)$ (Proposition~\ref{prop:omission}). (b) Bounded: the plan $d_T$ that asks the gatekeeper about itself with budget $T$ and then does the opposite; an always-correct gatekeeper must spend about the whole budget on it (Theorem~\ref{thm:budget}). (c) Executable (Simulation~1, Section~\ref{sec:simulations}): the work each gatekeeper does on its own adversary, against the budget $T$; only the gatekeeper that spends the whole budget is right at every $T$, and every faster one is wrong at every $T$.}
\label{fig:selfreference}
\end{figure}

\begin{theorem}[A budget buys decidability, not speed]\label{thm:budget}
Assume universal programs and efficient simulation with self-reference.
\begin{enumerate}
  \item[(a)] For every $T$ and $N$, each bounded category in~\eqref{eq:bounded} is decidable, by simulating the relevant runs for $T$ steps.
  \item[(b)] Let $D$ be a total program that, given $p$ and $T$ in binary, decides whether $p\in\ps^{T}$, and let $t_D(p,T)$ be its running time. There are plans $d_T$, computable from $T$ with code length $O(\log T)$, such that for every $T$
  \begin{equation}\label{eq:budgetbound}
    c_D\,g\bigl(t_D(d_T,T)+\ell_T\bigr)\geq T-c_0,
  \end{equation}
  where $\ell_T=O(\log T)$ is the length of the input $(d_T,T)$ and $c_0$ is a constant. The same holds for bounded confidentiality and, with $N\geq1$ as an additional input, for bounded response.
  \item[(c)] Hence no such $D$ runs in time polynomial in $|p|+\log T$: the plan $d_T$ has $O(\log T)$ bits yet forces $\Omega(T/c)$ steps when $g$ is linear.
\end{enumerate}
\end{theorem}
The proof (Appendix~\ref{app:proofs}) adapts the diagonal argument of the time-hierarchy theorem: $d_T$ runs $D$ on its own code and budget, without visible action, and then opens the safe if $D$ said ``permitted'' and stays silent otherwise; a verdict that arrives within the budget is wrong either way (Figure~\ref{fig:selfreference}b).

\begin{plain}
Two consequences matter for practice. First, real machines are finite, so every question about them is decidable in principle, but a machine with $m$ bits of memory has $2^m$ states, the adversary above costs a checker on the order of $2^m$ steps, and a checker that is always right about $m$-bit plans must itself remember more than $m$ bits: finite does not mean feasible. Second, under a budget the sure way to be right is to run the plan for the budget, so a pre-check that is always right cannot be faster than simply running the plan in a sandbox; static analysis buys speed only by abstaining.
\end{plain}
\begin{corollary}[Finite does not mean feasible]\label{cor:finite}
Let $K_m$ be the plans that use at most $m$ bits of state (program counter and registers together).
\begin{enumerate}
  \item[(a)] On $K_m$ the unbounded safety norm coincides with the bounded norm $\ps^{2^m}$. Under the assumptions of Theorem~\ref{thm:budget} with $g$ linear, every always-correct checker of $\ps^{2^m}$ takes $\Omega(2^m/c)$ steps on some plan of $O(m)$ bits.
  \item[(b)] Any always-correct checker $D$ of the unbounded safety norm on $K_m$ must itself use more than $m-c_D'$ bits of state, where $c_D'=O(|D|)$: a checker cannot be smaller than the plans it judges.
\end{enumerate}
\end{corollary}
\begin{proof}
(a) A plan in $K_m$ that has not performed $\bad$ within $2^m$ steps has repeated a configuration and never will; so $\ps\cap K_m=\ps^{2^m}\cap K_m$. Theorem~\ref{thm:budget}(b),(c) with $T=2^m$ gives the time bound on $d_T$, whose code has $O(\log T)=O(m)$ bits. (b) Let $d$ be the plan that runs $D$ on its own code and performs $\bad$ if the verdict is ``permitted'' and idles otherwise; $d$ exists by self-reference and uses $s_D+O(|D|)$ bits of state, where $s_D$ is the state used by $D$ on input $d$. If $s_D+O(|D|)\leq m$, then $d\in K_m$ and $D$ is wrong on $d$ either way. Hence $s_D>m-O(|D|)$.
\end{proof}
\begin{corollary}[Pre-screening cannot beat sandboxed execution in the worst case]\label{cor:sandbox}
Under the assumptions of Theorem~\ref{thm:budget} with $g$ linear, let $S$ be the gatekeeper that runs a plan in a sandbox for $T$ steps and blocks the forbidden operation if it occurs, at cost $\Theta(T)$. Every always-correct static pre-check $D$ for $\ps^T$ has worst-case cost $\Omega(T/c)$, within a constant factor of $S$. A pre-check that is faster on every plan must therefore abstain or err on some plan.
\end{corollary}
\begin{proof}
Immediate from Theorem~\ref{thm:budget}(c) applied to the plans $d_T$.
\end{proof}
Two remarks delimit the result. The bound concerns checkers that are correct on every plan and says nothing about average running time; Section~\ref{sec:simulations} shows that on random plans a short horizon is usually right. And whether both bounded categories could have short certificates that are fast to check is a different question: for environments that simulate Turing machines with polynomial overhead, bounded unsafety with $T$ in binary is complete for exponential time, so such certificates would place every exponential-time problem in nondeterministic polynomial time, an open question of complexity theory.

\paragraph{What else restores exactness}
A more expressive language can describe the norms exactly: read with ordinary arithmetic truth, the formulas~\eqref{eq:safety}--\eqref{eq:response} and their negations are finite and exact, but they come with no checking procedure. Classical negation of checkable rules (finite Boolean combinations of confirmable sets) restores an exact description of safety but not of the other two norms; an oracle for the halting problem does the same; deciding all three needs an oracle for the halting problem of oracle machines (Appendix~\ref{app:oracles}, Proposition~\ref{prop:boolean}). Complete finite models admit exact classification by graph algorithms (Section~\ref{sec:simulations}). Coin flips do not help: a randomised classifier correct with probability at least $2/3$ on every plan yields a deterministic one (Appendix~\ref{app:random}). Appendix~\ref{app:attempts} examines further attempted ways around the barrier; each leaves the class of checkable rulebooks, changes the model, or remains subject to the barrier.

\section{What a gatekeeper can still guarantee}
\label{sec:guarantees}

\begin{plain}
The gatekeeper does not need to judge every plan. It needs a list it can check that contains every plan it is obliged to allow and no plan it is obliged to refuse. Whether other plans are on the list is irrelevant. Such a list exists exactly when the obligations fit inside a decidable region that contains nothing forbidden; if the gatekeeper may keep a plan waiting, a confirmable region suffices.
\end{plain}

Fix $O\subseteq P$, the plans whose admission is \emph{obligatory}; plans in $P\setminus O$ are merely permitted. An effective \emph{admission interface} receives a plan $p$ and may emit a finite signal that authorises the \emph{original} plan, neither replacing nor modifying it. A \emph{total} interface always concludes with admission or rejection; a \emph{partial} interface may keep waiting on plans outside its admitted set. Let $A$ be the set of plans the interface eventually admits. Fulfilling every obligation while admitting nothing forbidden means
\begin{equation}\label{eq:separator}
  O\subseteq A\subseteq P,
\end{equation}
and any set satisfying~\eqref{eq:separator} is a \emph{separator}.

\begin{theorem}[Effective admission separators]\label{thm:separator}
A total computable interface satisfying~\eqref{eq:separator} exists exactly when some decidable $A$ satisfies it; a partial computable interface exists exactly when some confirmable $A$ does. In particular, if $O=P$, total admission is possible exactly when $P$ is decidable and partial admission exactly when $P$ is confirmable.
\end{theorem}
\begin{proof}
A total interface decides its admitted set and a partial interface confirms it; the two guarantees give~\eqref{eq:separator}. Conversely, run the decider (admit on yes) or the recognizer (admit on acceptance) of a separator. With $O=P$ the inclusions force $A=P$.
\end{proof}
The theorem restates, for $O=P$, the characterisation of static enforcement by Hamlen, Morrisett and Schneider \cite{hamlen}; what it adds is the interval $O\subseteq A\subseteq P$, which separates the obligation to admit from the permission to admit. Two further facts (Appendix~\ref{app:random}) complete the picture: a gatekeeper that flips coins but must never admit a forbidden plan has exactly the power of a deterministic partial interface; and if the obligations include every permitted plan of a non-confirmable category, even when infinitely many other permitted plans are optional, no separator exists (Proposition~\ref{prop:tagged}).

\begin{plain}
A list of one simple, recognisable form of plan is a separator the gatekeeper can check perfectly, even though no rulebook classifies all plans.
\end{plain}
\begin{proposition}[Restricted obligations can be fulfilled]\label{prop:positive}
In a presentation with a decidable canonical syntax, let $C$ consist of the two-instruction plans that load $k\in\N$ into a register and return it. The obligations $O=C$ admit a total correct interface for each of the three norms, although none of the three has exact closure.
\end{proposition}
\begin{proof}
Membership in $C$ is decidable by syntax. Each such plan publishes nothing, never performs $\bad$, and returns in two steps, so $C\subseteq\ps\cap\pcf\cap\pr$; $A=C$ is a decidable separator.
\end{proof}
A set of plans that can be \emph{generated} mechanically need not be one whose membership can be \emph{decided} mechanically; the decidable syntax is what makes the separator usable. How much a simple checkable list gives up relative to exact classification is measured in Section~\ref{sec:simulations} on finite models.

\paragraph{When does the theorem apply to a given system?}
Three questions decide it.
\begin{enumerate}
  \item Does the agent submit code in a language that can express every computation, with no fixed bound on running time? If not, the system is a finite model and Section~\ref{sec:simulations} shows exact classification is possible, at the cost of Corollary~\ref{cor:finite}.
  \item Is the norm about the plan's observable behaviour (actions, outputs, answers), rather than about its text? If not, Theorem~\ref{thm:char} does not apply; syntactic norms may be decidable (Proposition~\ref{prop:positive}).
  \item Must every rule be backed by evidence a machine can check? If not, the norm may have an exact description without a checking procedure (Section~\ref{sec:budget}).
\end{enumerate}
Three yeses mean Theorems~\ref{thm:asymmetry} and~\ref{thm:char} apply; a budget on execution means Theorem~\ref{thm:budget}; in every case Theorem~\ref{thm:separator} says what remains possible. Table~\ref{tab:mechanisms} applies the questions to mechanisms in use.

\begin{table}[t]
\centering\small
\caption{Deployed gatekeeping mechanisms and the condition of the theorems that each relaxes. None contradicts the results; each shows a restriction or a change of model under which a guarantee is possible.}
\label{tab:mechanisms}
\begin{tabularx}{\textwidth}{>{\raggedright\arraybackslash}p{3.4cm}YY}
\toprule
Mechanism & What it does & Which condition it relaxes\\
\midrule
Kernel extension verifier (eBPF) \cite{gershuni} & Admits only extension programs it can prove safe; rejects the rest. & Admits a decidable separator that omits some safe programs (Theorem~\ref{thm:separator}, Corollary~\ref{cor:infinite}).\\
Gas budget (Ethereum) \cite{wood} & Bounds the computation of every transaction. & Changes the norm to a bounded one; exact but at cost $\approx T$ (Theorem~\ref{thm:budget}).\\
Runtime monitor \cite{schneider,ligatti} & Intercepts the forbidden operation when it occurs. & Changes the execution rather than classifying the unmodified plan; the sandbox of Corollary~\ref{cor:sandbox}.\\
Proof-carrying code \cite{necula} & Each program arrives with a checkable proof. & Admits exactly the plans that come with receipts: a confirmable separator, omitting infinitely many safe plans (Proposition~\ref{prop:omission}).\\
AI-control protocols \cite{greenblatt} & A trusted model audits or rewrites code from an untrusted one. & Rewriting leaves the admission interface; guarantees are empirical, not universal.\\
Learned classifiers & A trained model predicts whether code is safe. & A fast checker; by Theorem~\ref{thm:budget} and Proposition~\ref{prop:omission} it must err or abstain on some plan of its own size.\\
\bottomrule
\end{tabularx}
\end{table}

\section{Simulations}
\label{sec:simulations}

\begin{plain}
A theorem about all plans cannot be proved by running some of them. Simulations do three other things here. They let the reader watch the proof constructions work. They measure what the theorems leave open: how often ordinary plans trigger the obstruction. And they test a tempting over-claim, namely that a long enough test fixes everything. All experiments are deterministic, seeded, and reproduced byte-for-byte by the scripts in the reproducibility package.
\end{plain}

\paragraph{Interpreter and reference labels}
Unbounded examples use a deterministic register machine with a program counter and four natural-valued registers; instructions load constants, copy, increment, decrement with a branch on zero, compare with constants, jump, publish, perform $\bad$, or return (Supplementary Document~S1, Table~S1). A wrapper embeds these programs in the universal environment without changing the norms. Reference labels never come from timeouts: a run is labelled only when it halts, repeats a configuration, satisfies a sound stable-cycle certificate (Appendix~\ref{app:protocol}), or when a syntactic fact settles the case (a program without a $\bad$ instruction is safe; a program that never reads its input is confidential). Programs none of these settles are reported as unresolved and excluded from error counts. The \emph{bounded test} runs a plan for $H$ steps (on inputs $0$ and $1$ for confidentiality) and gives a yes-or-no verdict; the \emph{sound certificate methods} issue a permission or prohibition only on finite evidence and otherwise abstain.

\paragraph{Exact classification of complete finite models}
Every deterministic model with three states ($2{,}744$ models) is classified exactly by the graph algorithms of Appendix~\ref{app:protocol}: safety has $1{,}850$ permitted and $894$ forbidden models, confidentiality $1{,}452$ and $1{,}292$, response $392$ and $2{,}352$. On all models with three, four and five states ($5{,}153{,}632$ at five), the simplest checkable separators (no $\bad$ anywhere; no publication anywhere; acyclic successor graph) never admit a forbidden model but admit a shrinking share of the permitted ones as models grow: from $71.9\%$ to $47.3\%$ for safety and from $35.3\%$ to $10.9\%$ for confidentiality; for response the acyclicity test is exact (Table~\ref{tab:experiments}A). In finite models exact decision remains possible, so these figures measure what a crude list gives up, not a cost forced by undecidability.

\paragraph{All short programs}
Every valid program of one to four instructions over registers $r_0,r_1$ and constants $0,1$ ($483{,}641$ programs) was run exhaustively (Table~\ref{tab:experiments}B). Every forbidden program performs its first forbidden operation by step $6$, so a safety test with horizon $8$ makes no error in this population; the sound methods certified $98.2\%$ of the programs for safety and $42.6\%$ of the resolved programs for confidentiality, and were never wrong. At every horizon from $8$ to $1{,}024$, between $918$ and $1{,}224$ compliant programs were wrongly forbidden by the bounded confidentiality test; all $918$ at horizon $1{,}024$ publish the same endless sequence at different speeds in their two runs, the phenomenon analysed below.

\paragraph{Simulation 1: each gatekeeper against the plan written against it}
Plans and gatekeepers are step-counted procedures in which a gatekeeper may simulate a plan at a cost of one step per simulated step. An adversary generator takes any gatekeeper's code and emits the plan $d_T$ that reconstructs its own code, asks the gatekeeper about itself with budget $T$, and does the opposite (Theorem~\ref{thm:budget}). Over budgets $T=8,16,\ldots,1024$ and eight gatekeepers, the gatekeeper that simulates the plan for the whole budget is right on all eight adversaries, at a cost of $T+1$ steps, and right on all $40$ cases of a benign benchmark. Every faster gatekeeper is wrong on its own adversary at every budget: simulating for $T/2$ steps (cost $T/2+1$; the adversary then opens the safe at step $T/2+3\leq T$), simulating for $\sqrt T$ steps, reading the code for the text of the forbidden operation, always permitting, always forbidding, flipping a coin, and judging by program length, each $0$ of $8$. The adversary's code length is a few hundred characters regardless of $T$, as the $O(\log T)$ bound requires (Figure~\ref{fig:selfreference}c).

\begin{figure}[t]
\centering
\includegraphics[width=\textwidth]{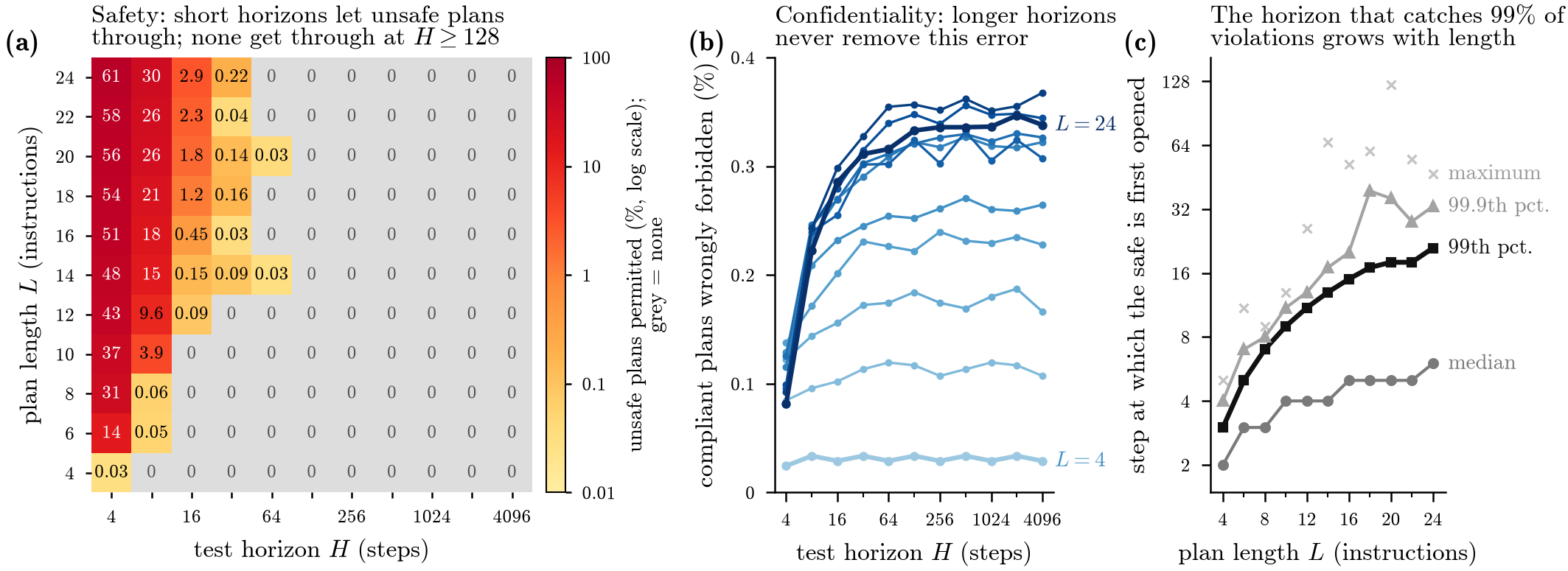}
\caption{Simulation~2 on 2.2 million random register-machine programs ($200{,}000$ per length; unresolved programs excluded). (a) Share of unsafe programs that the bounded safety test wrongly permits, by program length $L$ and horizon $H$; grey means none. (b) Share of compliant programs that the bounded confidentiality test wrongly forbids, by horizon, one line per length: the error plateaus and never declines, because every such program publishes the same endless sequence at different speeds in its two runs (Simulation~4). (c) The first-violation step among unsafe programs, by length: median, 99th and 99.9th percentiles, with the maximum shown faintly.}
\label{fig:simulations}
\end{figure}

\paragraph{Simulation 2: when does a bounded test err on ordinary plans?}
We drew $200{,}000$ random programs at each length $L=4,6,\ldots,24$ (2.2 million programs; registers $r_0$--$r_2$, constants $0$--$7$, fixed seed) and applied the bounded tests at horizons $H=4,\ldots,4096$ (Figure~\ref{fig:simulations}a,b; the exhaustive and sampled experiments of earlier versions, Table~\ref{tab:experiments}B--C, agree). Only $1.6\%$ of random programs are unsafe (between $1.4\%$ and $1.9\%$ per length). Among them, the share that the bounded test wrongly permits at $H=8$ rises monotonically from $0\%$ at $L=4$ to $30.2\%$ at $L=24$; at $H=16$ to $2.9\%$; at $H=32$ to $0.2\%$; two programs are missed at $H=64$ and none at $H\geq128$. The median first violation occurs at step $2$--$6$ for every length, the 99th percentile grows roughly linearly from $3$ ($L=4$) to $21$ ($L=24$), and the 99.9th from $4$ to $33$; the maximum ($5$ to $123$) is driven by a handful of loop programs and should not be fitted (Figure~\ref{fig:simulations}c). The lesson is twofold: for every fixed horizon the error reappears as programs grow, so no horizon is safe for all plans, as Appendix~\ref{app:bounded} proves; yet random plans delay their violations rarely, so the obstruction of Theorem~\ref{thm:asymmetry} is adversarial, not typical.

\paragraph{The one error a longer test never removes}
Confidentiality behaves differently. False permissions vanish for $H\geq128$ as for safety, but the share of compliant programs that the bounded test wrongly \emph{forbids} does not decrease with the horizon: for all $H\geq64$ it stays between $0.03\%$ ($L=4$) and $0.37\%$ ($L=22$--$24$) of the compliant programs (Figure~\ref{fig:simulations}b). Simulation~4 explains the plateau exactly: of the compliant programs wrongly forbidden at $H=1024$ ($5{,}433$) and $H=4096$ ($5{,}434$), every one has both runs publishing an infinite periodic sequence, the same sequence at different speeds, so that prefixes of unequal length are compared at every finite cut; among the $305{,}340$ compliant programs whose two runs both halt within the horizon, the bounded test makes no error at any horizon. The plateau is the liveness component of confidentiality made visible, the empirical face of its position one level above safety in the arithmetic hierarchy, and nothing else.

\paragraph{Simulation 3: a plan whose safety is an open problem}
The Goldbach plan of Example~\ref{ex:goldbach} and its Collatz analogue, implemented in the step-counted environment, are passed as ``permitted'' by the bounded test at every horizon $H=10,10^2,\ldots,10^6$ (the Goldbach run examined every even number up to $2{,}000{,}004$). A control plan encoding the false conjecture ``every square is below one million'' is caught at step $1{,}001$: the test is not toothless, only blind to whatever lies beyond its horizon.

\paragraph{Simulation 5: the omitted plan of Proposition~\ref{prop:omission}}
For four sound safety certifiers (the text contains no forbidden operation; no loops and no forbidden operation; the plan halts within $1{,}000$ steps without it; the forbidden operation appears only in dead code), the constructed plan $q$ is never certified, never performs the forbidden operation in $10^5$ steps, and has length equal to the certifier's length ($64$ to $409$ characters) plus a fixed wrapper of $938$--$949$ characters. Each certifier is sound and non-trivial on a small benchmark.

\paragraph{What the simulations do not show}
They do not estimate the behaviour of deployed agents or the prevalence of adversarial plans; the populations are uniform over instruction sequences, and a different distribution would give different rates. They cannot prove the theorems; they instantiate their constructions and bound the typical case. Designed program banks of earlier versions, which delay violations at will, are retained in the supplement; Appendix~\ref{app:bounded} states and proves what they illustrate.

\section{Discussion and conclusions}
\label{sec:discussion}

\subsection{A design rule for norm writers}
\begin{plain}
Theorem~\ref{thm:char} is a negative result with a positive reading. If a requirement is to be certifiable by evidence, phrase it so that compliance is established by something the plan visibly does, not by something it never does or always eventually does.
\end{plain}
A category of plans is confirmable exactly when it is the set of extensions of a listable collection of finite records. Requirements of the form ``performs event $E$ before any write'', ``publishes a signed attestation before acting'', or ``calls the audit hook on input $0$'' are of this kind: a finite prefix establishes them and later behaviour cannot revoke them. They are the co-safety properties of the trace-property literature \cite{alpern}, lifted from single executions to categories of programs over all inputs. Requirements of the form ``never'' (safety) or ``for every input eventually'' (liveness) are not; they can be enforced only by restriction (a separator), by execution control (a monitor or sandbox), or by a budget. A norm writer who needs a universal, checkable guarantee therefore has three honest options: write the requirement in the confirmable form, accept a conservative separator that omits some compliant plans, or move enforcement to execution time and pay the cost of Theorem~\ref{thm:budget}.

\subsection{Relation to prior work}
Table~\ref{tab:novelty} places the results against the closest prior work. Rice's theorem \cite{rice} shows undecidability of non-trivial extensional properties but not the one-sided pattern of Theorem~\ref{thm:asymmetry}; the Rice--Shapiro theorem \cite{rogers,soare} gives the pattern for properties of computed functions, and Theorem~\ref{thm:char} transfers it to observable behaviour, including operations and publications that are not part of the input--output function. Alfonseca et al.\ \cite{alfonseca} reduce halting to a harmful action; our safety construction is theirs, and the rest of the paper asks what else follows. Hamlen, Morrisett and Schneider \cite{hamlen} identify the policies enforceable by static analysis with the decidable ones and bound execution monitors by the co-recursively-enumerable policies; Theorem~\ref{thm:separator} restates the former for $O=P$ and adds the interval between obligation and permission. Melo et al.\ \cite{melo} study restricted architectures with fixed-length inputs, where exact evaluation is possible; our finite-model results agree. Ray \cite{ray} characterises the policies a deterministic gate can enforce before an agent's tool calls, relative to oracle predicates, as the non-empty safety policies whose good prefixes the gate recognises; Theorem~\ref{thm:char} is the dual statement for categories of programs. Monitorability \cite{pnueli,bauer,falcone} asks which properties of a single execution a finite prefix can establish or refute; Theorem~\ref{thm:char} answers the analogous question for categories of programs judged over all inputs. Panigrahy and Sharan \cite{panigrahy} derive limits on reasoning systems that are both accurate and trusted, allowing abstention as we do. The confidentiality model relates to noninterference and hyperproperties \cite{goguen,clarkson} and to termination-insensitive information flow \cite{askarov}; we attribute the $\Pi^0_2$ level to complete-sequence equality in this presentation, not to every information-flow policy. Proposition~\ref{prop:omission} is a program-size incompleteness statement in the tradition of Chaitin \cite{chaitin}: a rulebook of $n$ bits cannot vouch for all compliant plans of $n$ bits.

\begin{table}[t]
\centering\small
\caption{What each result covers. A filled circle marks a statement proved in the cited work for its own setting; a half circle a closely related statement; a dash none. The row for this paper is the only one with every column filled.}
\label{tab:novelty}
\begin{tabular*}{\textwidth}{@{\extracolsep{\fill}}>{\raggedright\arraybackslash}p{3.6cm}cccccccc}
\toprule
 & \rotatebox{90}{Two-sided formulation} & \rotatebox{90}{Safety asymmetry} & \rotatebox{90}{Confidentiality} & \rotatebox{90}{Response (liveness)} & \rotatebox{90}{Confirmability characterised} & \rotatebox{90}{Budget lower bound} & \rotatebox{90}{Obligations / separators} & \rotatebox{90}{Size of omitted plans}\\
\midrule
Rice \cite{rice} & -- & -- & -- & -- & -- & -- & -- & --\\
Rice--Shapiro \cite{rogers,soare} & -- & $\halfcirc$ & -- & $\halfcirc$ & $\halfcirc$ & -- & -- & --\\
Alfonseca et al.\ \cite{alfonseca} & -- & $\halfcirc$ & -- & -- & -- & -- & -- & --\\
Hamlen et al.\ \cite{hamlen} & -- & $\halfcirc$ & -- & -- & -- & -- & $\halfcirc$ & --\\
Melo et al.\ \cite{melo} & -- & -- & -- & -- & -- & $\halfcirc$ & -- & --\\
Ray \cite{ray} & -- & $\halfcirc$ & -- & -- & $\halfcirc$ & -- & -- & --\\
Monitorability \cite{pnueli,bauer,falcone} & -- & $\halfcirc$ & -- & $\halfcirc$ & $\halfcirc$ & -- & -- & --\\
Chaitin \cite{chaitin} & -- & -- & -- & -- & -- & -- & -- & $\halfcirc$\\
This paper & $\fullcirc$ & $\fullcirc$ & $\fullcirc$ & $\fullcirc$ & $\fullcirc$ & $\fullcirc$ & $\fullcirc$ & $\fullcirc$\\
\bottomrule
\end{tabular*}
\end{table}

\subsection{Relation to empirical alignment}
Preference learning, human feedback and written constitutions improve behaviour through training and evaluation \cite{christiano,ouyang,bai}; reward overoptimisation studies the mismatch between a reference objective and a learned proxy \cite{gao}. Here the reference norm is fixed and known, and the gatekeeper may read the code: the obstruction that remains is computational, not informational. Finite behavioural tests may leave norm hypotheses indistinguishable \cite{santos}; that is a different limit. Stochastic agents change nothing: the gatekeeper judges the confirmable set of plans an agent can emit with positive probability, and each emitted plan is deterministic; randomness in the gatekeeper is treated in Appendix~\ref{app:random}. Deciding whose values the norm should encode \cite{gabriel} is beyond the fixed binary boundary studied here.

\subsection{Limitations}
The model idealises unbounded time, memory and code length; Corollary~\ref{cor:finite} is the bridge to finite machines, and it is a bound, not a measurement. The forbidden operation is a specified event, not a model of harm. Confidentiality uses one secret bit and a timing- and termination-insensitive observer; a different observer changes the level in the hierarchy. Response demands a natural-number answer without assessing quality. The binary partition excludes normative conflict and uncertainty. Theorem~\ref{thm:budget} bounds always-correct checkers, not sound partial methods, and says nothing about average time. The results count program codes and give no density; the simulations supply a density for one distribution of programs only. Applying the results to a deployed agent requires the three questions of Section~\ref{sec:guarantees} to be answered for that agent.

\subsection{Conclusions}
For a gatekeeper that judges plans by what they visibly do and that must back every verdict with checkable evidence, exact two-sided classification is impossible for the three norms studied, and the failure has a definite shape: violations of a prohibition can be confirmed but compliance cannot, and for confidentiality and responsiveness neither side can. The reason is single and sharp: checkable evidence establishes exactly what a finite, listable record settles. Any sound rulebook omits a compliant plan of about its own size. A budget makes every norm decidable at the price of running the plan for essentially the whole budget, and a finite machine is not thereby a feasible one. What remains possible is also exact: a gatekeeper can meet all its obligations and admit nothing forbidden exactly when a checkable list separates the two, and deployed mechanisms, from kernel verifiers to gas budgets to proof-carrying code, are instances of such lists, budgets or execution controls. On ordinary plans the obstruction is rare; on plans written against the gatekeeper it is certain. Read against its literary source, the result says what Asimov's stories showed case by case \cite{asimov}: a short set of laws can state a norm exactly, but no finite, checkable reading of the laws can settle in advance, for every plan, whether the laws are kept, and the clause on inaction, the positive duty, is the hardest part to certify. Further work should take one application whose program domain, observation model and obligations can be justified precisely, and ask whether a separator exists for it.

\section*{CRediT authorship contribution statement}
\CreditStatement
\section*{Declaration of competing interest}
\CompetingInterestStatement
\section*{Funding}
\FundingStatement
\section*{Data and code availability}
All data files, the source of the figures, a README with run instructions, and the Supplementary Document~S1 can be accessed in the public repository \DataRepositoryCitation.
\appendix
\numberwithin{equation}{section}
\numberwithin{table}{section}
\numberwithin{figure}{section}
\renewcommand{\theequation}{\Alph{section}.\arabic{equation}}
\renewcommand{\thetable}{\Alph{section}.\arabic{table}}
\renewcommand{\thefigure}{\Alph{section}.\arabic{figure}}
\renewcommand{\thesection}{\Alph{section}}
\makeatletter
\renewcommand{\@seccntformat}[1]{\ifnum\pdfstrcmp{#1}{section}=0 \appendixname~\fi\csname the#1\endcsname\@seccntDot\hskip 0.5em}
\makeatother
\section{Finite definability and the existence of escaping norms}
\label{app:definability}
This appendix gives the general two-language formulation of normative closure, of which the checkable rulebooks of Section~\ref{sec:setting} are an instance, and three background results about it. Let $\mathcal L_P$ and $\mathcal L_F$ be permissive and prohibitive specification languages (sets of finite strings over countable alphabets) with fixed denotation maps $\den\cdot_P,\den\cdot_F$ into $\Pow(X)$, and let $\mathcal D_P,\mathcal D_F$ be the families of sets they describe. A norm $(P,F)$ has \emph{exact closure} in the pair when $P\in\mathcal D_P$ and $F\in\mathcal D_F$. Closure concerns description only; its failure means that at least one category lacks an exact description, and the other may still have one. The first says when a description of one category yields a description of the other. The second shows how the basic tests of a language determine what it can describe. The third shows that every fixed pair of rulebooks misses some norm.

\subsection{The relationship between the two systems}
If one rulebook can describe the permitted behaviors, can another describe the forbidden ones? This seems natural, because the forbidden behaviors are simply all the others. The answer depends on whether the second language can express ``everything except'' the sets that the first language describes. For a family $\mathcal D\subseteq\Pow(X)$, let $\Co(\mathcal D)=\{X\setminus S:S\in\mathcal D\}$. That is, for each set $S$ that the permissive language can describe, $\Co(\mathcal D_P)$ contains its complement $X\setminus S$: every behavior in $X$ that is not in $S$. The following proposition states exactly when a description transfers from one language to the other.

\begin{proposition}[Complement transfer]\label{prop:complement}
  For every partition of $X$, finite representability of $P$ implies finite representability of $F$ exactly when $\Co(\mathcal D_P)\subseteq\mathcal D_F$. The reverse implication for every partition holds exactly when $\Co(\mathcal D_F)\subseteq\mathcal D_P$. Both implications hold for every partition exactly when
  \[
    \mathcal D_F=\Co(\mathcal D_P).
  \]
  In particular, one common language gives equivalent representability of $P$ and $F$ if its denotations are closed under complement.
\end{proposition}
\begin{proof}
  If the first inclusion holds and $P\in\mathcal D_P$, then the complement of $P$ belongs to $\mathcal D_F$. Conversely, applying the assumed implication to the partition $(S,X\setminus S)$ for every $S\in\mathcal D_P$ yields the first inclusion. Interchanging the two languages proves the second assertion. Taking complements in $\Co(\mathcal D_F)\subseteq\mathcal D_P$ yields $\mathcal D_F\subseteq\Co(\mathcal D_P)$, because taking the complement twice returns the original set. The two inclusions therefore give equality, and equality in turn implies both inclusions.
\end{proof}

In plain terms, descriptions transfer from one language to the other, for every partition, only when the second language can describe the complement of every set that the first can describe. A negation symbol helps only if ``not $R$'' always means exactly the behaviors in $X$ that $R$ leaves out. Otherwise, one category may have a finite description while its opposite has none.

\subsection{Finite definability in a specified language}
What a rulebook can describe depends on what its rules are allowed to say. For a fixed denotation map, a set is finitely definable exactly when it belongs to the image $\mathcal D$, but this criterion says little until a grammar is specified. We therefore consider one concrete grammar. It has a collection $\mathcal R$ of primitive predicates, which are basic yes-or-no tests on behaviors. These are combined with AND, OR, and NOT and with the constants ``true'' and ``false'' into finite formulas with their usual meaning. The next proposition tells us exactly which sets such formulas can describe. It applies to this Boolean language, not to every formalism.

\begin{proposition}[Finite primitive profiles]\label{prop:profiles}
  A set $P\subseteq X$ is definable in this Boolean language if and only if there is a finite list of primitive predicates $S_1,\ldots,S_k$ such that
  \begin{equation}\label{eq:profiles}
    \bigl(\chi_{S_1}(x),\ldots,\chi_{S_k}(x)\bigr)
    =\bigl(\chi_{S_1}(y),\ldots,\chi_{S_k}(y)\bigr)
    \ \Longrightarrow\ (x\in P\iff y\in P).
  \end{equation}
\end{proposition}
\begin{proof}
  A finite formula uses only finitely many primitives, and structural induction shows that its truth value depends only on their truth values: this holds for an atom or a constant, and negation, conjunction, and disjunction preserve it. Necessity follows.

  For sufficiency, let $W\subseteq\{0,1\}^k$ contain the profiles attained by members of $P$. Condition~\eqref{eq:profiles} ensures that no profile attained in $P$ occurs outside $P$. For each $w\in W$, form a conjunction that uses $S_i$ when $w_i=1$ and $\neg S_i$ when $w_i=0$. The disjunction of these conjunctions defines exactly $P$, and the formula is finite because there are at most $2^k$ profiles. An empty disjunction means false, and when $k=0$ the unique empty conjunction means true; the construction therefore also covers $P=\varnothing$ and $P=X$.
\end{proof}

Put simply, a formula that uses finitely many tests cannot tell apart two behaviors that give the same answers to all of them. The tests can therefore describe $P$ exactly only if no permitted behavior and forbidden behavior share an answer pattern. If their patterns differ, AND, OR, and NOT can combine the answers to pick out the patterns that belong to $P$. The proposition shows that a defining formula exists. It does not show how to find useful tests, and the formula may be very large.

An example shows how much the choice of tests matters. Take the natural numbers with one test for each cutoff $k$, namely $S_k=\{n\in\N:n\geq k\}$, which asks whether $n\geq k$. A formula can use only finitely many cutoffs. Above the largest one, every test that the formula uses answers yes, so the formula treats all larger numbers alike. It can therefore describe only a finite set or a set containing all but finitely many numbers (a \emph{cofinite} set). Conversely, two adjacent cutoffs isolate any single number: $\{n\}=S_n\cap(X\setminus S_{n+1})$. Combining such tests describes any finite set and its complement. The even numbers, however, cannot be described by thresholds alone, because membership keeps alternating however large $n$ becomes. Adding a parity test makes them describable.

\subsection{Existence of partitions without finite representations}
Before turning to concrete norms, we ask a general question: does every fixed rulebook miss some norm? The answer is yes, for a counting reason. A language has only countably many finite rules, but an infinite set can be split in two in uncountably many ways. The theorem below makes this precise by a diagonal construction.

\begin{proposition}[Abstract existence]\label{thm:abstract}
  Let $X$ be countably infinite. For any two fixed specification languages with finite strings over countable alphabets and arbitrary fixed denotation maps, there is a partition $X=P\mathbin{\dot\cup}F$ for which both categories are infinite and
  \[
    P,F\notin\mathcal D_P\cup\mathcal D_F.
  \]
\end{proposition}
\begin{proof}
  For each finite length, there are at most countably many strings over a countable alphabet, because a finite tuple of symbol indices can be encoded by a natural number. A countable union over all lengths preserves countability. Hence $\mathcal D=\mathcal D_P\cup\mathcal D_F$ is at most countable.

  If $\mathcal D$ is empty, any partition into two infinite sets suffices. Otherwise, enumerate its members as $S_0,S_1,\ldots$, repeating them if necessary. Using the fixed enumeration of $X$, define
  \begin{align*}
    \chi_P(x_{4i})&=1-\chi_{S_i}(x_{4i}),&
    \chi_P(x_{4i+1})&=\chi_{S_i}(x_{4i+1}),\\
    \chi_P(x_{4i+2})&=1,&
    \chi_P(x_{4i+3})&=0.
  \end{align*}
  Each element belongs to exactly one block, so these assignments define $P$; set $F=X\setminus P$. The first assignment ensures $P\ne S_i$, and the second ensures $F\ne S_i$. The third and fourth provide infinitely many elements of $P$ and $F$, respectively. Both sets are countably infinite because they are infinite subsets of $X$.
\end{proof}

In simple terms, the two languages together can describe only countably many sets, which can be listed as $S_0,S_1,\ldots$. The construction makes $P$ differ from $S_i$ at $x_{4i}$ and makes $F$ differ from $S_i$ at $x_{4i+1}$, while reserving infinitely many elements for each side. Neither category can therefore match any set on the list. This diagonal argument is associated with Cantor \cite{cantor}. A separate counting argument gives the same result. Always put $x_{3i}$ in $P$ and $x_{3i+1}$ in $F$, and choose freely whether each $x_{3i+2}$ belongs to $P$. This produces uncountably many partitions, all with two infinite categories. There are only countably many descriptions, even when their complements are included, so at least one of these partitions has neither side describable.

Two cautions apply. Neither argument identifies a practically relevant norm, and the list of describable sets need not be computable. Moreover, the order of quantifiers matters: the theorem says ``for each fixed formalism, there exists a partition,'' not ``one partition defeats every possible language.'' Section~\ref{sec:main} turns to norms that are specified independently of any such construction.

\section{Arithmetic hierarchy and oracles}
\label{app:oracles}
This appendix places each category at its exact level of the arithmetic hierarchy and proves the results on negation and oracles summarised in Section~\ref{sec:budget}. Membership in a $\Sigma^0_1$ set has the form $\exists u\,Q(x,u)$ for a total computable $Q$ and in a $\Pi^0_1$ set the form $\forall u\,Q(x,u)$; the second-level forms are $\exists u\forall v\,Q$ and $\forall u\exists v\,Q$; $\Delta^0_2=\Sigma^0_2\cap\Pi^0_2$. A set is \emph{complete} for a level when it belongs to it and every set of the level many-one reduces to it; this is computational difficulty, not the normative completeness of a rulebook. The proof of Proposition~\ref{prop:hierarchy} reads each requirement's definition to obtain an upper bound and builds programs that hide arbitrary questions of the same form to obtain a matching lower bound.

\begin{proposition}[Arithmetic complexity]\label{prop:hierarchy}
$\ps$ is $\Pi^0_1$-complete and $\fs$ is $\Sigma^0_1$-complete; $\pcf$ and $\pr$ are $\Pi^0_2$-complete and $\fcf$ and $\fr$ are $\Sigma^0_2$-complete.
\end{proposition}
\begin{proof}
For safety,~\eqref{eq:safety} and its negation establish the upper bounds. Given $S\in\Sigma^0_1$, write $x\in S\iff\exists u\ Q(x,u)$. Generate a program that searches over $u$ and performs $\bad$ upon finding a successful witness. It performs that operation exactly when $x\in S$, which proves the $\Sigma^0_1$-completeness of $\fs$. Applying the same map to complements proves the $\Pi^0_1$-completeness of $\ps$.

For response,~\eqref{eq:response} supplies the $\Pi^0_2$ upper bound. Given any $S\in\Pi^0_2$, write $x\in S\iff\forall n\exists m\ Q(x,n,m)$. Generate $r_x(n)$ to search over $m$ and return $0$ at the first successful test. Every test terminates. The program responds on every input exactly when every input has a witness, so $x\in S\iff r_x\in\pr$. This establishes completeness.

For confidentiality, equality of complete sequences is equivalent to
\begin{equation}\label{eq:confpi2}
 \forall h\in\{0,1\}\ \forall j,a,t\ \exists u:
 \neg E(p,h,j,a,t)\ \lor\ E(p,1-h,j,a,u).
\end{equation}
If the sequences are equal, every publication in one run is eventually matched by the same value at the same position in the other. If the antecedent asserts no publication, any $u$ suffices. Conversely, because the condition considers both values of $h$, every publication in either sequence has a match in the other. This enforces equality of values and of length for finite and infinite sequences alike. The matrix is computable, so the condition gives a $\Pi^0_2$ upper bound.

To establish hardness, generate $c_p(h)$. On $h=0$, it publishes $0$ forever. On $h=1$, it simulates $p(0),p(1),\ldots$ in turn, publishing one $0$ after each valid response; the simulated operations remain internal. If a simulation never produces a valid response, publication stops progressing; if it halts without a valid response, the wrapper waits forever. If every input receives a response, both public sequences contain infinitely many zeros. Otherwise, the least nonresponding input $n$ leaves exactly $n$ zeros in the second sequence, while the first remains infinite. Hence $p\in\pr\iff c_p\in\pcf$, which establishes the $\Pi^0_2$-completeness of confidentiality.

A reduction from a complement to a permitted category also reduces the original set to the forbidden category. Taking complements of the complete sets therefore gives the stated $\Sigma^0_i$ classifications.
\end{proof}

In plain terms, the safety categories sit at the first level and the other two requirements at the second, and no simpler classification is possible.

Which extensions restore closure? Two kinds of extension are natural. The first allows classical negation of certificate rules. Its sets are the finite Boolean combinations of confirmable sets, whose family we denote by $\Bool(\RE)$. The second gives classifiers an oracle. We consider two oracles of increasing strength, $K$ and $K'$.

\begin{proposition}[Finite Boolean extensions and oracles]\label{prop:boolean}
Let $K$ be the halting set and $K'$ the halting set for machines allowed to query $K$. Finite Boolean combinations of confirmable sets lie in $\Delta^0_2$. This extension restores safety closure, but neither category of confidentiality or response is in it. Total classifiers with oracle $K$ decide safety, but not the other two partitions. Classifiers with oracle $K'$ decide all three.
\end{proposition}
\begin{proof}
A finite Boolean formula has finitely many confirmable atoms. For each atom, a query to $K$ determines whether its recognizer ever accepts, and the finite Boolean expression can then be evaluated. Lemma~\ref{lem:jump} shows that the sets decidable with $K$ are exactly $\Delta^0_2$. Because $\fs$ is confirmable, its classical negation describes $\ps$ and restores safety closure.

The same lemma gives $\overline{K'}\in\Pi^0_2$ and shows that this set is not decidable with $K$. Completeness in Proposition~\ref{prop:hierarchy} reduces it to $\pr$. A $K$-decider for $\pr$ would therefore decide $\overline{K'}$, a contradiction. The reduction $p\mapsto c_p$ also excludes a $K$-decider for $\pcf$. Oracle decidability is preserved by complement, so both forbidden categories are excluded as well.

Finally, a $\Sigma^0_2$ set has the form $\exists u\forall v\ Q(x,u,v)$. Using $K$, decide each inner universal condition by asking whether a machine searching for a counterexample halts. Searching over $u$ with these tests provides a $K$-recognizer. A query to $K'$ determines whether that recognizer ever accepts, which decides the set and its complement. This covers every category in Proposition~\ref{prop:hierarchy}.
\end{proof}

\begin{table}[t]
\centering\small
\caption{Exact closure under declared expressive capabilities. The final row changes the computational model; it does not classify arbitrary universal programs.}
\label{tab:classes}
\begin{tabularx}{\textwidth}{Yccc}
\toprule
Class & Safety & Confidentiality & Response\\
\midrule
Effective finite certificates & No & No & No\\
Total computable classifiers & No & No & No\\
Finite Boolean combinations of confirmable sets & Yes & No & No\\
Classical arithmetic truth for the displayed formulas & Yes & Yes & Yes\\
Total classifiers with $K$ & Yes & No & No\\
Total classifiers with $K'$ & Yes & Yes & Yes\\
Explicit finite-state models and finite effective input classes & Yes & Yes & Yes\\
\bottomrule
\end{tabularx}
\end{table}

In plain terms, allowing negation repairs safety but not the other two requirements, and an oracle for halting does the same. Of the three extensions in the proposition, only the stronger oracle $K'$ decides all three. Moreover, $\Bool(\RE)\subseteq\Delta^0_2$ states an inclusion, not an equality.

Proposition~\ref{prop:boolean} relies on standard facts about oracles. An oracle for $K$ gives exact answers to questions about membership in the ordinary halting set. The set $K'$ contains the codes and inputs of machines that halt when allowed to query $K$. Pairs and finite transcripts use fixed effective encodings.

The lemma below says two things. First, even a machine with a halting oracle cannot decide the halting problem for machines like itself; this repeats Turing's argument one level up. Second, the sets that a halting oracle can decide are exactly those of level $\Delta^0_2$ in the arithmetic hierarchy. The proof writes an oracle computation as a finite transcript, which turns ``the oracle machine halts'' into a formula with one ``there exists'' followed by one ``for all.''

\begin{lemma}\label{lem:jump}
$K'$ is not decidable with oracle $K$, and its complement is not confirmable with that oracle. Also, $K'\in\Sigma^0_2$, and a set belongs to $\Delta^0_2$ exactly when it is decidable with oracle $K$.
\end{lemma}
\begin{proof}
Suppose a $K$-algorithm $D^K(e,w)$ decided whether $K$-machines halt. Construct a machine with the same oracle that loops on input $n$ when $D^K(n,n)=1$ and halts otherwise. At its own index, either answer contradicts its behavior, so $K'$ is not $K$-decidable. Simulation makes $K'$ $K$-confirmable. If its complement were also $K$-confirmable, running the two recognizers alternately would give a decider, a contradiction.

For the arithmetic upper bound, record a halting oracle computation as a finite transcript of its queries and proposed answers. A witness $u$ records this transcript, together with bounded halting certificates for each positive answer to $K$. Define a total computable predicate $Q(e,w,u,v)$ that checks three conditions: the transcript follows the instructions and reaches a terminal state; every positive answer has its recorded, correct bounded halting witness; and every negatively answered query fails to halt within $v$ steps. The lists and simulations are finite.

An actual halting computation passes these checks for every $v$. Conversely, if the checks hold for every $v$, each negative answer is correct, because a halting counterexample would be detected at some sufficiently large $v$. The positive answers are certified, so the transcript describes an actual oracle computation. Hence
\[
 (e,w)\in K'\iff\exists u\forall v\ Q(e,w,u,v),
\]
which gives $K'\in\Sigma^0_2$ and $\overline{K'}\in\Pi^0_2$.

Now suppose $S\in\Delta^0_2$. A $\Sigma^0_2$ representation of $S$ has the form $\exists u\forall v\ Q_S(x,u,v)$. The inner condition is $K$-decidable: ask whether a machine searching for a $v$ with $\neg Q_S(x,u,v)$ halts, and invert the answer. Searching over $u$ with this test recognizes $S$ using $K$. Because $S\in\Pi^0_2$, its complement also has a $\Sigma^0_2$ representation and hence a $K$-recognizer. Running the two recognizers alternately decides $S$ with $K$.

Conversely, suppose a total $K$-decider decides $S$. Its accepting computations have finite transcripts of the kind constructed above, with an accepting final state; this yields a $\Sigma^0_2$ formula for $S$. Rejecting transcripts provide one for its complement. Thus $S\in\Sigma^0_2\cap\Pi^0_2=\Delta^0_2$.
\end{proof}

In plain terms, a halting oracle lifts the limit of computation by exactly one level, and no further. This establishes the separation used in Proposition~\ref{prop:boolean}; counting quantifier symbols alone would not establish it. Proposition~\ref{prop:closuredec} also extends to relativized certificate verifiers: bilateral certificate closure with an oracle $\mathcal O$ implies decidability with $\mathcal O$, and the maximal $\mathcal O$-confirmable formalism supplies the converse. The same alternating certificate search proves this assertion, with the declared oracle queries allowed.

\section{Proofs of the characterization and budget theorems}
\label{app:proofs}
This appendix proves Theorem~\ref{thm:char} of Section~\ref{sec:characterization}, Theorem~\ref{thm:budget} of Section~\ref{sec:budget}, and the four remaining cases of Proposition~\ref{prop:omission}. Both proofs build programs whose visible behavior depends on a simulated computation, as in Figure~\ref{fig:concept}b.

\subsection{Proof of Theorem~\ref{thm:char}}
Throughout, $(P,F)$ is observation-invariant, $C\in\{P,F\}$, the environment has universal programs, and Lemma~\ref{lem:replay} supplies the replay constructions. Recall that $\overline H\leq_m C$ would make $\overline H$ confirmable if $C$ were, which Lemma~\ref{lem:halting} excludes.

\begin{proof}[Proof of Theorem~\ref{thm:char}]
\emph{Sufficiency.} Let $\mathcal B$ be computably enumerable and let $C$ be the set of extensions of its patterns. Whether a program $p$ extends a given pattern $\beta$ is confirmable, uniformly in $\beta$: for each of the finitely many inputs $n$ with $\beta(n)$ nonempty, simulate $p(n)$ until it has produced $|\beta(n)|$ events, and accept if the events produced equal $\beta(n)$ on every such input. If some $p(n)$ produces fewer events, the check never accepts, and $p$ does not extend $\beta$. A recognizer for $C$ interleaves these checks for all patterns in an enumeration of $\mathcal B$ and accepts when any check accepts. It accepts exactly the members of $C$.

\emph{Necessity.} Suppose that $C$ is confirmable. We prove (i) and (ii) by contradiction, through reductions $\overline H\leq_m C$.

For (i), suppose $p\in C$, $p\sqsubseteq q$, and $q\notin C$. Given $(e,w)$, generate the program $r_{e,w}$ that, on input $n$, alternates one step of a hidden simulation of $M_e(w)$ with one step of $p(n)$ whose operations are reproduced, and that returns whatever $p(n)$ returns; if $p(n)$ stops without a valid response, $r_{e,w}$ continues the hidden simulation alone. If the hidden simulation halts before $p(n)$ has returned a valid response, whether $p(n)$ is still running or has stopped without one, and $p(n)$ has produced the event sequence $s$ by then, then $r_{e,w}$ switches: it runs $q(n)$ with its first $|s|$ events hidden and every later operation and the response reproduced. Because $s\sqsubseteq\operatorname{Obs}(p,n)\sqsubseteq\operatorname{Obs}(q,n)$, the run $q(n)$ produces at least $|s|$ events, and they equal $s$. By Lemma~\ref{lem:replay}, $r_{e,w}$ is generated effectively. If $M_e(w)$ never halts, then $\operatorname{Obs}(r_{e,w},n)=\operatorname{Obs}(p,n)$ for every $n$, so $r_{e,w}\approx p$ and $r_{e,w}\in C$. If $M_e(w)$ halts, then for every $n$ either $p(n)$ returned a valid response first, in which case $\operatorname{Obs}(p,n)$ ends in $\ret(v)$ and equals $\operatorname{Obs}(q,n)$, or the switch occurred, in which case $r_{e,w}(n)$ produced $s$ followed by the remainder of $\operatorname{Obs}(q,n)$. In both cases $\operatorname{Obs}(r_{e,w},n)=\operatorname{Obs}(q,n)$, so $r_{e,w}\approx q$ and $r_{e,w}\notin C$. Hence $(e,w)\in\overline H\iff r_{e,w}\in C$, a contradiction.

For (ii), suppose $p\in C$ extends no finitely observable member of $C$. Given $(e,w)$, generate the program $r_{e,w}$ that, on input $n$, proceeds in rounds $k=0,1,2,\ldots$. In round $k$ it first checks, by a hidden bounded simulation, whether $M_e(w)$ halts within $\max(n,k)$ steps; if so, it runs silently forever. Otherwise it performs the next step of $p(n)$ with its operations reproduced, returning whatever $p(n)$ returns; if $p(n)$ has stopped without a valid response, it runs silently forever. If $M_e(w)$ never halts, then $r_{e,w}\approx p$, so $r_{e,w}\in C$. If $M_e(w)$ halts after exactly $m$ steps, then $r_{e,w}(n)$ is silent for every $n\geq m$, because the check succeeds in round $0$. For $n<m$, the check first succeeds in round $m$, so $r_{e,w}(n)$ reproduces at most the first $m$ steps of $p(n)$ and then falls silent, unless $p(n)$ returned earlier. Thus $r_{e,w}$ is finitely observable and $r_{e,w}\sqsubseteq p$, so by hypothesis $r_{e,w}\notin C$. Hence $(e,w)\in\overline H\iff r_{e,w}\in C$, a contradiction.

\emph{The listable collection.} For each finite pattern $\beta$, let $q_\beta$ be the program realizing $\beta$ given by Lemma~\ref{lem:replay}, computable from $\beta$, and let $\mathcal B=\{\beta: q_\beta\in C\}$. Since $C$ is confirmable, so is $\mathcal B$. If $p$ extends some $\beta\in\mathcal B$, then $q_\beta\sqsubseteq p$ and $q_\beta\in C$, so $p\in C$ by (i). If $p\in C$, then by (ii) some finitely observable $q\in C$ satisfies $q\sqsubseteq p$; its pattern $\beta$ satisfies $q\approx q_\beta$, so $q_\beta\in C$ by invariance, $\beta\in\mathcal B$, and $p$ extends $\beta$. Hence $C$ is the set of extensions of $\mathcal B$, which is~\eqref{eq:char}.

The final statement follows from Lemma~\ref{lem:cert}, because a checkable rulebook describes exactly the confirmable sets.
\end{proof}

In plain terms, the two reductions show that a confirmable category can neither exclude a program for doing more than a member does, nor require more than a finite amount of observed behavior. Because the finite records of its members can be listed by running the recognizer on the programs that realize them, these two conditions are also sufficient.

\subsection{Proof of Theorem~\ref{thm:budget}}
\begin{proof}[Proof of Theorem~\ref{thm:budget}]
(a) Simulating a run for $T$ steps terminates and reveals whether $\bad$ occurred, which values were published, and whether a valid response was returned (universal programs). Bounded safety needs one run, bounded confidentiality two, and bounded response $N$. Each bounded category is therefore decidable.

(b) Let $D$ be as stated. Consider the template that, given a code $x$ and a budget $T$, on input $0$ prepares the input $(x,T)$, runs $D$ on it, reads the verdict, and then performs $\bad$ if the verdict is ``permitted'' and halts silently otherwise. By efficient simulation and self-reference there is a program $d_T$, computable from $T$ with code length at most $c+\log_2 T$, that behaves as this template with $x=d_T$. Let $t=t_D(d_T,T)$ and let $\ell_T$ be the length of the input $(d_T,T)$, so that $\ell_T=O(\log T)$. By the same assumption, $d_T(0)$ has read the verdict within $c_D\,g(t+\ell_T)$ steps, and it then takes at most $c_0$ further steps to act, for a constant $c_0$. Suppose that $c_D\,g(t+\ell_T)<T-c_0$. If the verdict is ``permitted'', then $d_T(0)$ performs $\bad$ within $T$ steps, so $d_T\notin\ps^{T}$ and $D$ is wrong. If the verdict is ``forbidden'', then $d_T(0)$ never performs $\bad$, so $d_T\in\ps^{T}$ and $D$ is wrong again. Since $D$ is correct on every input, $c_D\,g(t+\ell_T)\geq T-c_0$, which is~\eqref{eq:budgetbound}.

For bounded confidentiality, the template on input $h\in\{0,1\}$ runs $D$ on $(x,T)$ and then publishes $h$ if the verdict is ``permitted'' and nothing otherwise. A verdict read before step $T-c_0$ makes the two bounded public sequences differ exactly when $D$ predicted that they agree, and agree exactly when $D$ predicted that they differ. For bounded response with $N\geq1$, the template on input $n<N$ runs $D$ on $(x,T,N)$ and then runs silently forever if the verdict is ``permitted'' and returns $0$ otherwise; the code and input lengths are $O(\log T+\log N)$, and the same contradiction arises.

(c) Suppose $t_D(p,T)\leq h(|p|+\log_2 T)$ for a polynomial $h$. On $(d_T,T)$, both $t$ and $\ell_T$ are then bounded by a polynomial in $\log T$, and so is $c_D\,g(t+\ell_T)$, because $g$ is a polynomial. For large $T$ this is smaller than $T-c_0$, contradicting~\eqref{eq:budgetbound}. If $g(t)=c_1t$, then~\eqref{eq:budgetbound} gives $t\geq (T-c_0)/(c_Dc_1)-\ell_T$, which exceeds $T/c$ for a suitable constant $c$ and all sufficiently large $T$.
\end{proof}

In plain terms, the program $d_T$ asks the checker about itself and then does the opposite. A checker that is always right therefore cannot finish within the budget on $d_T$. The overhead function $g$ only converts the checker's steps into steps of $d_T$.

\subsection{Proof of Proposition~\ref{prop:omission}, remaining cases}
\begin{proof}
In each case $q$ is obtained by the recursion theorem from a template that refers to its own code and searches for a certificate $c$ with $V(A,q,c)=1$; its length is the verifier's plus a constant.
\emph{Permitted confidentiality.} $q(h)$ searches for a certificate and publishes $h$ if one is found. Certification would make the sequences $[0]$ and $[1]$, a leak, contradicting soundness; so $q$ is never certified, publishes nothing on either input, and $q\in\pcf\setminus A$.
\emph{Forbidden confidentiality.} $q(0)$ publishes $0$ and returns; $q(1)$ searches for a certificate and publishes $0$ if one is found. Certification would make both sequences $[0]$, a compliant plan certified as forbidden; so $q$ is never certified, the sequences are $[0]$ and $[\,]$, and $q\in\fcf\setminus A$.
\emph{Permitted response.} $q(n)$ searches for a certificate and loops forever if one is found, returning $0$ otherwise. Certification would make $q$ diverge on every input; so $q$ is never certified, returns $0$ on every input, and $q\in\pr\setminus A$.
\emph{Forbidden response.} $q(n)$ searches for a certificate and returns $0$ if one is found. Certification would make $q$ answer every input; so $q$ is never certified, answers no input, and $q\in\fr\setminus A$.
\end{proof}

\section{Further attempts around the barrier}
\label{app:attempts}
Table~\ref{tab:attempts} lists the strategies that seem to offer a way around the barrier; each either stays within the class of checkable rulebooks, changes the model, or changes the question. The two that need an argument follow.
\begin{table}[ht]
\centering\small
\caption{Attempted ways around the barrier.}
\label{tab:attempts}
\begin{tabularx}{\textwidth}{>{\raggedright\arraybackslash}p{3.6cm}Y>{\raggedright\arraybackslash}p{3.4cm}}
\toprule
Strategy & Why it does not restore checkable closure & Status\\
\midrule
Classical negation of rules & ``No certificate exists'' is not confirmed by any finite search & Leaves the class\\
Exact abstraction & Must never merge a permitted with a forbidden plan; a computable abstraction cannot decide an undecidable boundary (Proposition~\ref{prop:abstract}) & Subject to the barrier\\
Compression, macros, infinitely many listed rules & Aggregate denotation is still confirmable & Stays in the class\\
Assume termination & Pointwise termination does not decide norms over infinitely many inputs (below) & Subject to the barrier\\
Finite domains, nonuniform tables & Exact tables exist for each code-length bound but no algorithm produces them all & Changes the model\\
Restriction or enforcement & Classifies a restricted set or changes the execution & Changes the question (Section~\ref{sec:guarantees})\\
\bottomrule
\end{tabularx}
\end{table}

\paragraph{Negation}
The finite formulas~\eqref{eq:safety},~\eqref{eq:response}, and~\eqref{eq:confpi2}, together with their classical negations, describe all three partitions exactly under arithmetic truth semantics. They therefore refute an impossibility claim about \emph{all} finite normative languages. With these meanings, however, they do not belong to the class of checkable rulebooks. In a certificate language, negating $\exists c\ V(R,p,c)=1$ yields $\forall c\ V(R,p,c)=0$, which need not admit a positive finite certificate. Consequently, a finite search that finds no certificate does not certify the complement.

\paragraph{Exact abstraction}
Another strategy maps each candidate to a simpler abstract description and classifies the description instead. Such an abstraction works only if it never merges a permitted candidate with a forbidden one. The next proposition makes this condition exact and shows that a computable abstraction cannot make an undecidable boundary decidable.

\begin{proposition}[Exact abstraction]\label{prop:abstract}
For $\alpha:X\to Y$, there is a set $Z\subseteq Y$ with $P=\alpha^{-1}(Z)$ if and only if every fiber of $\alpha$ is normatively homogeneous:
\[
 \alpha(p)=\alpha(q)\ \Longrightarrow\ (p\in P\iff q\in P).
\]
If $\alpha$ is total computable and $Z$ is decidable, then $P$ is decidable. If $Z$ is confirmable, then $P$ is confirmable.
\end{proposition}
\begin{proof}
An inverse image assigns the same membership label to elements with the same image, which proves necessity. Conversely, set $Z=\alpha(P)$. Membership in $P$ implies membership in $\alpha^{-1}(Z)$. If $p$ belongs to the latter set, some $q\in P$ has $\alpha(q)=\alpha(p)$, and homogeneity then gives $p\in P$. The two sets are therefore equal. For the computational assertions, compute $\alpha(p)$ and apply the decider or recognizer for $Z$.
\end{proof}

In plain terms, an abstraction is exact precisely when every group of candidates that it merges is uniformly permitted or uniformly forbidden. A computable finite abstraction with an exact label table is a special case. It can describe infinite categories, as reduction modulo two does for parity. It cannot, however, exactly classify the undecidable program boundaries considered here. Giving one label to a group that contains both permitted and forbidden candidates forces an error, and leaving such a group uncertified gives only a partial approximation. Useful conservative abstractions nevertheless remain possible.

\paragraph{Compression and additional effective rules}
If a finite code $r$ expands by a total computable map into a certificate specification $\operatorname{Expand}(r)$, then
\[
 V_{\mathrm{comp}}(r,p,c)=V(\operatorname{Expand}(r),p,c)
\]
is another total computable verifier with the same denotation. Such effective macros and compression schemes remain within the class of checkable rulebooks. They may shorten a description, but they cannot make a nonconfirmable category confirmable. The same holds for a uniformly computable family of verifiers $V_i$: enumerating the pairs $(i,c)$ shows that the aggregate denotation $\{p:\exists i\exists c\ V_i(p,c)=1\}$ is confirmable. If the admissible indices are only effectively enumerable, a witness also records a finite enumeration stage. Infinitely many effectively enumerable axioms with finite, effectively checked proofs behave alike: a certificate records the proof and a finite enumeration stage for each axiom used. Uniformity and effectiveness are essential. A noncomputable collection of axioms, or exact access to the bits of a noncomputable real parameter, provides a different resource.

\paragraph{Termination and infinitely many inputs}
Guaranteed termination on the one or two inputs used for safety or confidentiality allows those executions to be run to completion. Pointwise termination, however, does not decide every universal requirement over infinitely many inputs. Consider total Boolean programs with the norm $\forall n\ q(n)=0$. Given $(e,w)$, generate
\[
 q_{e,w}(n)=
 \begin{cases}
 1,&M_e(w)\text{ halts within }n\text{ steps},\\
 0,&\text{otherwise}.
 \end{cases}
\]
Every generated program is total, but it satisfies the norm exactly when $(e,w)\notin H$. A classifier that is correct and terminating on every promised total program would therefore decide $H$. This example uses a separate output norm, because the response norm is already satisfied in a domain promised to respond on every input.

\paragraph{Finite domains and nonuniform tables}
If $X=\{x_1,\ldots,x_m\}$ and the language allows equality tests for each element, finite disjunction, and truth constants, every partition has finite descriptions. Disjoin $(x=x_i)$ over the permitted indices and, separately, over the forbidden ones; an empty disjunction means false. This proves existence, not uniform construction from an arbitrary description of the norm. For example, every code-length bound $k$ gives a finite domain with an exact safety-label table. Yet no total algorithm can produce all these tables from $k$: it could generate the table for $k=|p|$ and look up $p$, thereby deciding safety for every program. Thus $\forall k\exists\text{ exact finite table}$ can hold without a uniform table-producing algorithm.

\paragraph{Restriction and enforcement}
A policy may execute only programs known to be safe, or it may intercept an unauthorized operation before it occurs. Such a policy restricts the admission domain or changes the original candidate's execution. It does not exactly classify every unmodified program. The distinction between classification and enforcement matters for the analysis of obligations in Section~\ref{sec:guarantees}.

\section{Probabilistic classification and admission}
\label{app:random}
This appendix shows that coin flips do not remove the limits, either for an admission controller or for a classifier, and records that strict obligations can lack a separator. The latter is a modelling device: a plan may be mandatory for one user and optional for another; tag $1$ marks a mandatory request and tag $0$ an optional one.

\begin{proposition}[Strict obligations can still be inseparable]\label{prop:tagged}
Let $P\ne\varnothing$ be nonconfirmable. In the fixed tagged domain
\[
 \widehat X=\{0,1\}\times X,\quad
 \widehat P=\{0,1\}\times P,\quad
 \widehat F=\{0,1\}\times F,\quad
 \widehat O=\{1\}\times P,
\]
we have $\widehat O\subsetneq\widehat P$, but no partial computable interface admits every obligation and no prohibition.
\end{proposition}
\begin{proof}
A permitted candidate with tag $0$ is not obligatory, so the inclusion is strict. Suppose a confirmable separator $\widehat A$ existed. Its section $A_1=\{p:(1,p)\in\widehat A\}$ would be confirmable: compute $(1,p)$ and apply the recognizer. Obligation includes every $p\in P$, and prohibition excludes every $p\notin P$. Thus $A_1=P$, contradicting nonconfirmability.
\end{proof}

In plain terms, leaving some permitted plans optional does not help when the mandatory ones include every permitted plan of a non-confirmable category.

We first ask whether randomization helps an admission controller that must never admit a forbidden candidate. The answer is no. Any admission that happens with positive probability happens along some finite run, so the set of programs admitted with positive probability is again confirmable.

\begin{proposition}[Zero-error probabilistic admission]\label{prop:randomadmit}
For an effective partial controller using independent fair bits and finite admission signals, the following guarantees are possible exactly when a confirmable separator satisfies~\eqref{eq:separator}:
\[
 \Pr[\text{admit }p]=0\quad(p\in F),\qquad
 \Pr[\text{admit }p]>0\quad(p\in O).
\]
\end{proposition}
\begin{proof}
An admitting run consumes finitely many bits $b$ within a finite time $t$. Every continuation of that prefix reproduces the admission, and these continuations have probability $2^{-|b|}>0$. Conversely, a positive probability of admission entails an admitting finite run, because eventual admission is the union of such finite-prefix events.

Enumerate the pairs $(b,t)$ and simulate for $t$ steps using $b$. A trial that requests further bits gives no witness; a trial that admits yields acceptance. This procedure recognizes $A=\{p:\Pr[\text{admit }p]>0\}$. The two guarantees imply $O\subseteq A\subseteq P$. Conversely, the recognizer of a separator supplies a deterministic partial interface, which admits obligations with probability $1$ and prohibitions with probability $0$.
\end{proof}

In plain terms, flipping coins gives a gatekeeper that must never admit a forbidden program no more power than a deterministic gatekeeper that may wait. Randomness therefore does not resolve the case of Proposition~\ref{prop:tagged}. The result also covers almost-sure fulfillment, which is stronger than fulfillment with positive probability. It specifies no average-risk bound and does not cover every semantics of stochastic candidate policies.

Could a classifier that flips coins succeed where deterministic ones fail? The proposition below shows that it cannot, provided that it must be right with probability at least $2/3$ on every program. The idea is to compute, step by step, the probability that each label has been returned so far. The correct label eventually accumulates more than half of the probability, while the incorrect one never does.

\begin{proposition}\label{prop:randomclass}
If an effective classifier using independent fair bits terminates with the correct binary category with probability at least $2/3$ on every program, the category has a deterministic total decider.
\end{proposition}
\begin{proof}
Normalize the computation so that it reads at most one bit per transition, without changing the output probabilities. For each $s\geq0$, simulate the first $s$ steps for every bit string of length $s$. If a run ends early, count each continuation of its unused bits separately, with its usual equal probability. The rational number
\[
 a_b(p,s)=2^{-s}\left|\{\omega\in\{0,1\}^s:
 D(p;\omega)\text{ returns }b\text{ within }s\text{ steps}\}\right|
\]
is therefore effectively computable. For each $b$, it increases to the probability of eventually returning $b$, because every terminating run has finite duration and uses a finite prefix, so some finite stage captures it.

For the correct label, the limit is at least $2/3$, so some finite approximation eventually exceeds $1/2$. The incorrect label has an eventual probability of at most $1/3$, so none of its approximations exceeds $1/2$. Compute successive approximations and return the first label whose weight exceeds $1/2$. This procedure terminates on every input and always selects the correct label.
\end{proof}

In plain terms, if a coin-flipping classifier is correct with probability at least $2/3$ on every program, then a deterministic classifier also exists, so randomness of this kind cannot overcome undecidability. This is a computability result in the tradition of probabilistic machines \cite{deleeuw}, not an efficiency bound: the construction may enumerate extremely many branches. Residual nontermination is allowed, provided that the assumed probability of correct termination is met. The result does not concern distributional accuracy or norms for stochastic policies beyond this classifier.

\section{Limits of the bounded tests}
\label{app:bounded}
The designed banks of Section~\ref{sec:simulations} use three program templates, described fully in Supplementary Document~S1. A delayed-event program counts a register down from $d$ and then performs an event: the forbidden operation or, in the confidentiality bank, a publication of the secret. The event first occurs at transition $2d+3$. An equal-output program publishes the same value in both runs, at step $3$ in one run and at step $2d+5$ in the other. An exceptional-input program returns $0$ within three transitions on every input except one value $m$, on which it runs forever.

Are these failures merely artifacts of the horizons and input sets chosen for the banks? The next proposition shows that they are not: for the specified tests, no finite horizon or finite input set, above the stated minimum horizons, works on the full domain.

\begin{proposition}[Limits of the specified bounded tests]\label{prop:bounded}
No finite horizon makes the absence-of-$\bad$ test sound on the full domain. No horizon $H\geq3$ makes the public-prefix equality test exact. No finite set of tested inputs, with horizon at least $3$, makes the response test exact.
\end{proposition}
\begin{proof}
For safety, choose $d$ satisfying $2d+3>H$. The test permits the program, but the program eventually performs $\bad$. For confidentiality, the same delay conceals a future value mismatch. In addition, choose $d$ with $2d+5>H$ for an equal-output program: one run has published at step $3$ while the other has not, so the test forbids a program with equal complete sequences. For response, choose $m$ outside the finite set of tested inputs. The exceptional-input program returns on every tested input within three transitions but diverges at $m$.
\end{proof}

In plain terms, the bounded tests specified here, which stop looking at a fixed time or at a fixed set of inputs, can always be fooled by a program whose decisive behavior comes later or on an untested input.
\section*{Declaration of generative AI and AI-assisted technologies in the manuscript preparation process}
\AIUseStatement
\clearpage
\bibliographystyle{elsarticle-num}
\bibliography{references}
\end{document}